\documentclass[%
 aip,
 amsmath,amssymb,
 reprint,%
]{revtex4-1}

\usepackage{graphicx}
\usepackage{dcolumn}
\usepackage{bm}
\usepackage[utf8]{inputenc}
\usepackage[T1]{fontenc}
\usepackage{mathptmx}
\usepackage{etoolbox}
\usepackage{subfig}
\usepackage{hyperref}

\makeatletter
\def\@email#1#2{%
 \endgroup
 \patchcmd{\titleblock@produce}
  {\frontmatter@RRAPformat}
  {\frontmatter@RRAPformat{\produce@RRAP{*#1\href{mailto:#2}{#2}}}\frontmatter@RRAPformat}
  {}{}
}%
\makeatother
\begin{document}

\preprint{AIP/123-QED}

\title{Improving agent performance in fluid environments by perceptual pretraining}
\author{Jin Zhang, Jianyang Xue, Bochao Cao$^*$}
 \email{cbc@fudan.edu.cn}
\affiliation{ 
Department of Aeronautics and Astronautics, Fudan University, Shanghai 200433, China}

\begin{abstract}
In this paper, we construct a pretraining framework for fluid environment perception, which includes an information compression model and the corresponding pretraining method. We test this framework in a two-cylinder problem through numerical simulation. The results show that after unsupervised pretraining with this framework, the intelligent agent can acquire key features of surrounding fluid environment, thereby adapting more quickly and effectively to subsequent multi-scenario tasks. In our research, these tasks include perceiving the position of the upstream obstacle and actively avoiding shedding vortices in the flow field to achieve drag reduction. Better performance of the pretrained agent is discussed in the sensitivity analysis.
\end{abstract}

\maketitle

\section{Introduction}
In water environments, aquatic animals or underwater robots can be regarded as embodied intelligent agents who collect information (pressure, temperature, vision, etc.) from their surrounding water environments using their sensory organs or electronic sensors. And these time-series information would then be processed by these underwater agents and help them make flow field perception or fulfill underwater missions, such as cruising, obstacle avoidance, and so on.

Embodied intelligent agents acquire knowledge through interactions with their environment, often with limited information, and apply this knowledge across various tasks \cite{SmithLinda2005TDoE,DuanJiafei2022ASoE}. Reinforcement learning (RL) has emerged as a promising technique for training embodied agents in complex fluid environments\cite[e.g.][]{BruntonStevenL2020MLfF, VignonC2023Raia}. In the framework of RL, agents update their decision-making strategies through trials and errors, and finally maximize their cumulative rewards. There have been many successful cases of training embodied agents using RL algorithms. For example, RL has been adopted for training gliders to improve gliding performance in complex convection environments \cite[e.g.][]{ReddyGautam2018Gsvr, NovatiGuido2019Cgap}, for active flow control to reduce drag \cite[e.g.][]{RabaultJean2019Annt, FanDixia2020Rlfb}, and for studying the behavior of both individual fish-like swimmers and fish schooling from a biomimetic perspective \cite[e.g.][]{GazzolaMattia2014RLaW, VermaSiddhartha2018Ecsb, ZhuYi2021Anso}. Research works by Paris et al. \cite{ParisRomain2021Rfca} and \citet{LiJichao2022Rcoc} indicate the existence of optimal sensor placement for specific scenarios. Placing sensors in highly sensitive regions enables effective flow control even with a limited number of detection points. Wang et al. \cite{WangQiulei2024Dfdr} propose that integrating temporal data into agent state variable can significantly enhance performance of the agent. These findings reveal potential advantages of flow control strategies using deep reinforcement learning techniques.

However, most of previous studies focus on training agents to perform well in isolated tasks within fixed scenarios. Given the inherent complexity and nonlinearity of fluid dynamics, training agents to generalize their abilities across different fluid environment remains extremely challenging which demands innovative approaches in both RL frameworks and training methodologies.

Extraction of meaningful representations from collected data might be crucial before training agents to fulfill multiple tasks. Some recent research works have indicated that unsupervised pretraining allows agents to develop a deep understanding of complex patterns and relationships within the data, enhancing their ability to generalize from one task to another. This procedure has been validated in various areas, including large language models \cite[e.g.][]{DevlinJacob2019BPoD, LiJunnan2023BBLP}, computer vision \cite[e.g.][]{ChenTing2020ASFf,he2022masked,gao2024mind3dreconstructhighquality3d} and reinforcement learning in video game environments \cite[e.g.][]{SrinivasAravind2020CCUR, StookeAdam2021DRLf}. In the area of fluid dynamics, information compression method has also been introduced in several research works. For instance, Fukami et al. \cite{FukamiKai2019Srot} and Murata et al. \cite{MurataTakaaki2020Nmdw} use nonlinear encoders to compress spatial features of flow fields into low-dimensional manifolds. Racca et al. \cite{RaccaAlberto2023Ptdw} compress flow field data from both temporal and spatial dimensions using nonlinear encoders and successfully predict the development of unsteady turbulent flows by compressed low-dimensional information. 

In this study, we build a spatiotemporal compression model to transform perceptual information collected from a fluid environment into actionable knowledge. We start with unsupervised pretraining which enables the agent to learn environmental information without data labeling. Subsequently, using pretraining knowledge, we test the ability of the agent in different tasks, including obstacle position prediction and drag reduction reinforcement learning. The results illustrate that perceptual compression significantly enhances agent performance across different tasks. Our approach provides a new way of training general-purpose agents capable of various tasks in complex fluid environments.

\begin{figure}
  \centering
  \subfloat[]{
    \includegraphics[width=0.45\textwidth]{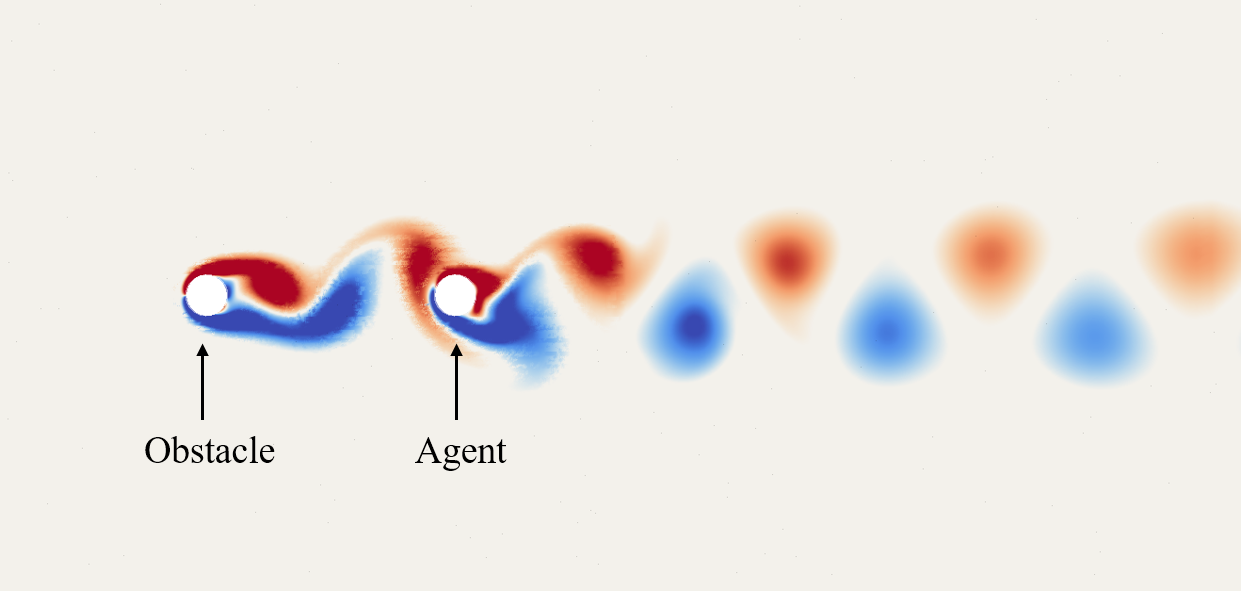}
    \label{fig1:subfig1}
  }
  \hfill
  \subfloat[]{
    \includegraphics[width=0.45\textwidth]{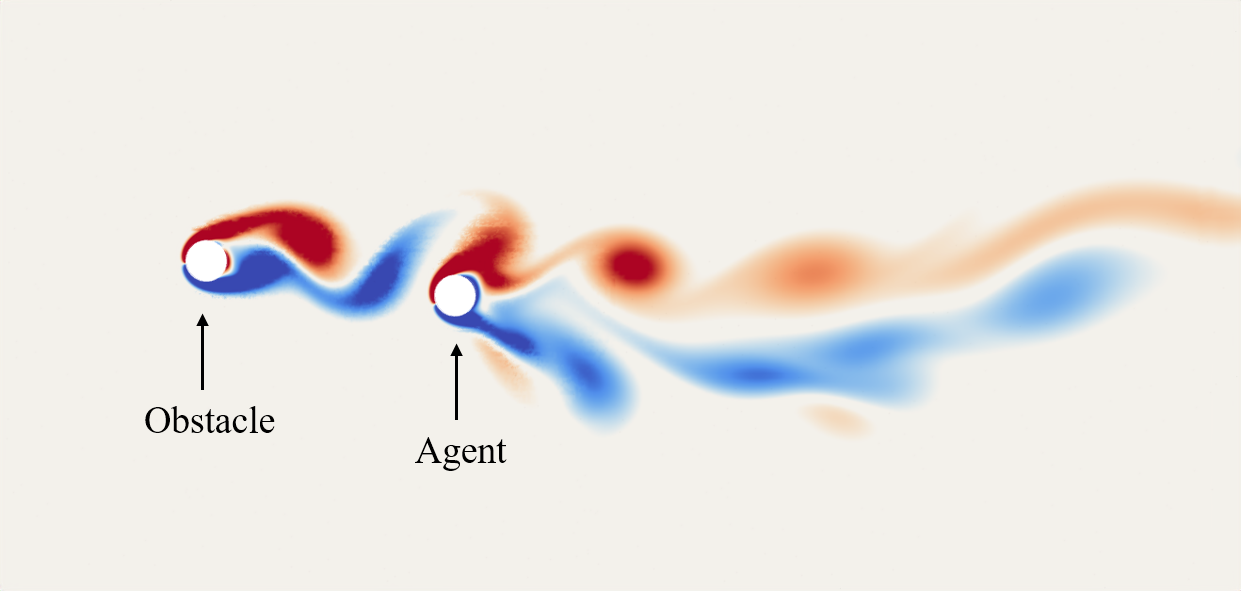}
    \label{fig1:subfig2}
  }
  \caption{Vorticity contours of the two-cylinder system. (a) Fixed obstacle cylinder; (b) Obstacle cylinder oscillating in the vertical direction.}
  \label{fig:1}
\end{figure}

\section{Methodology}
\subsection{Numerical Method}
In this research, flow around two streamwisely aligned identical cylinders is simulated in OpenFOAM (\href{http://www.openfoam.com}{www.openfoam.com}). The downstream cylinder acts as the perceptual intelligent agent in our study, while the upstream one serves as the obstacle which could oscillate in the vertical direction as shown in figure \ref{fig:1}. The Reynolds number of the incoming flow is set as 100, and hence laminar algorithm is used in the simulation and current problem is solved by incompressible Navier-Stokes equations as shown below, 
\begin{equation}
\frac{\partial u_i}{\partial x_i} = 0
\end{equation}

\begin{equation}
\frac{\partial u_i}{\partial t} + u_j \frac{\partial u_i}{\partial x_j} = -\frac{1}{\rho} \frac{\partial p}{\partial x_i} + \nu \frac{\partial^2 u_i}{\partial x_j \partial x_j}
\end{equation}

where \( u_i \) denotes velocity component in the \( i \)-th direction, \( \rho \) is the fluid density, \( p \) is the pressure, and \( \nu \) is the kinematic viscosity.

\begin{figure}
  \centering
  \subfloat[]{
    \includegraphics[width=0.45\textwidth]{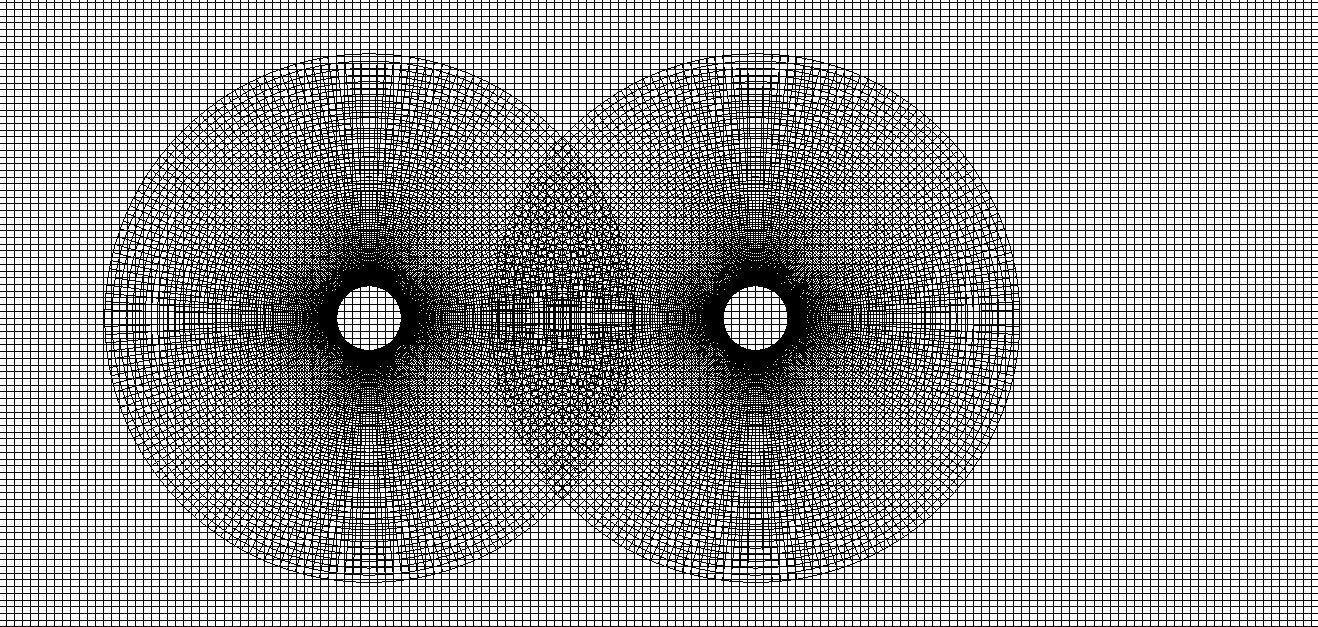}
    \label{fig2:subfig1}
  }
  \hfill
  \subfloat[]{
    \includegraphics[width=0.45\textwidth]{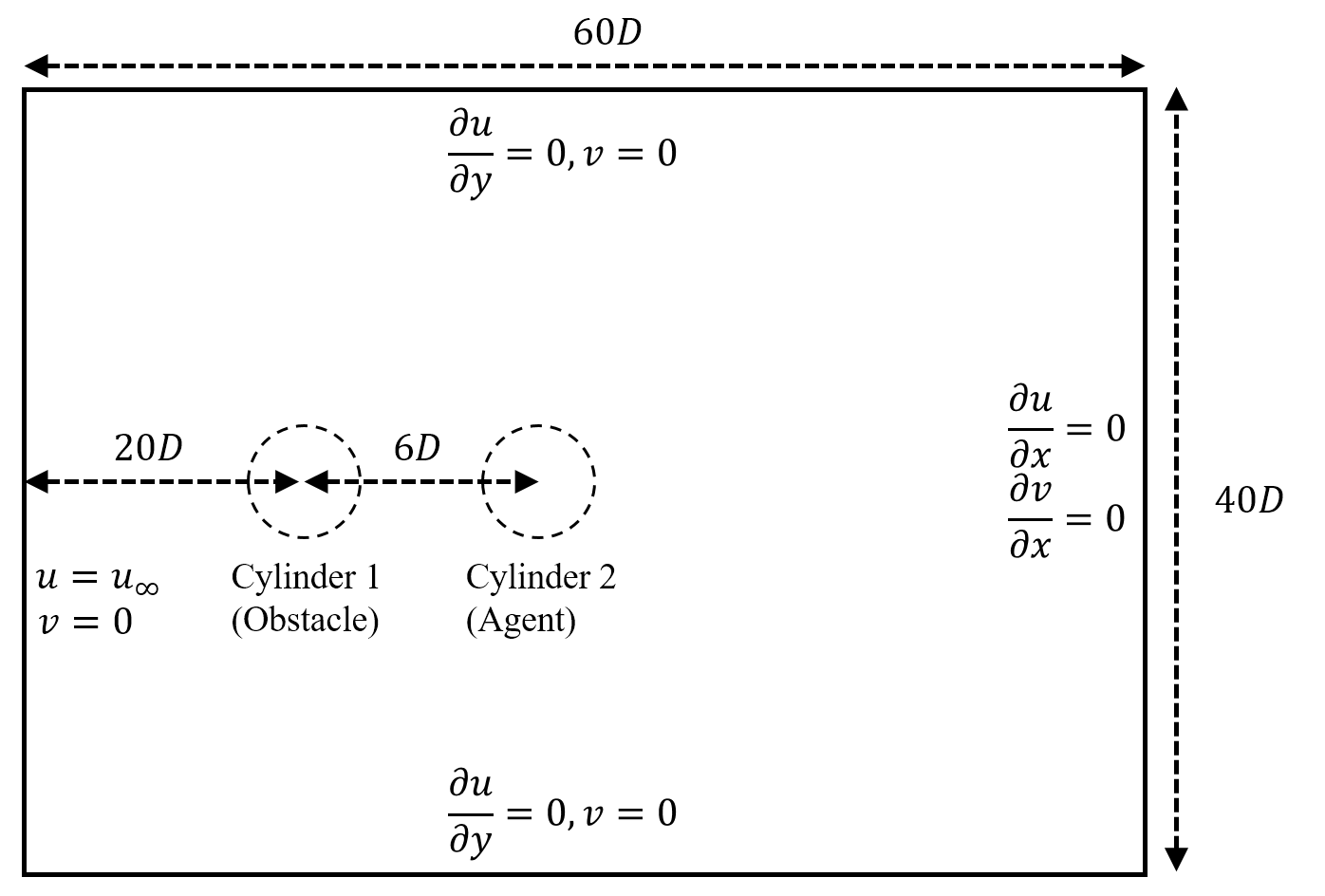}
    \label{fig2:subfig2}
  }
  \caption{Computational domain and grid setup. (a) Detailed view of the overset mesh around the cylinders. (b) Computational domain and boundary conditions.}
  \label{fig:2}
\end{figure}

In the simulation, using overset mesh technique, the computational domain is partitioned into overlapping subdomains, each with its own grid. A background mesh spans the entire computational domain, while refined grids are used around the cylinders, as shown in figure \ref{fig2:subfig1}. Computational information is exchanged between these grids through interpolation in overlapping regions. The illustration of computational domain is shown in figure \ref{fig2:subfig2}. The background mesh size is 60D x 40D, where D is the cylinder diameter. The mesh around each cylinder is set in a circular region with diameter of 8D. The cylinders are placed along the centerline of the computational domain, with cylinder 1 located 20D from the inlet and cylinder 2 6D downstream of cylinder 1. Free-stream condition is given at the inlet of computational domain, while far-field condition is set at the other three boundaries, as shown in figure \ref{fig2:subfig2}. A no-slip boundary condition is imposed on the cylinder surfaces. Moreover, since cylinder 2 is set as the perceptual agent in this study, its surface pressure is collected at each computational time step. Validation of current computational method is provided in APPENDIX \ref{Appendix A}.

\begin{figure*}
\includegraphics[width=0.8\textwidth]{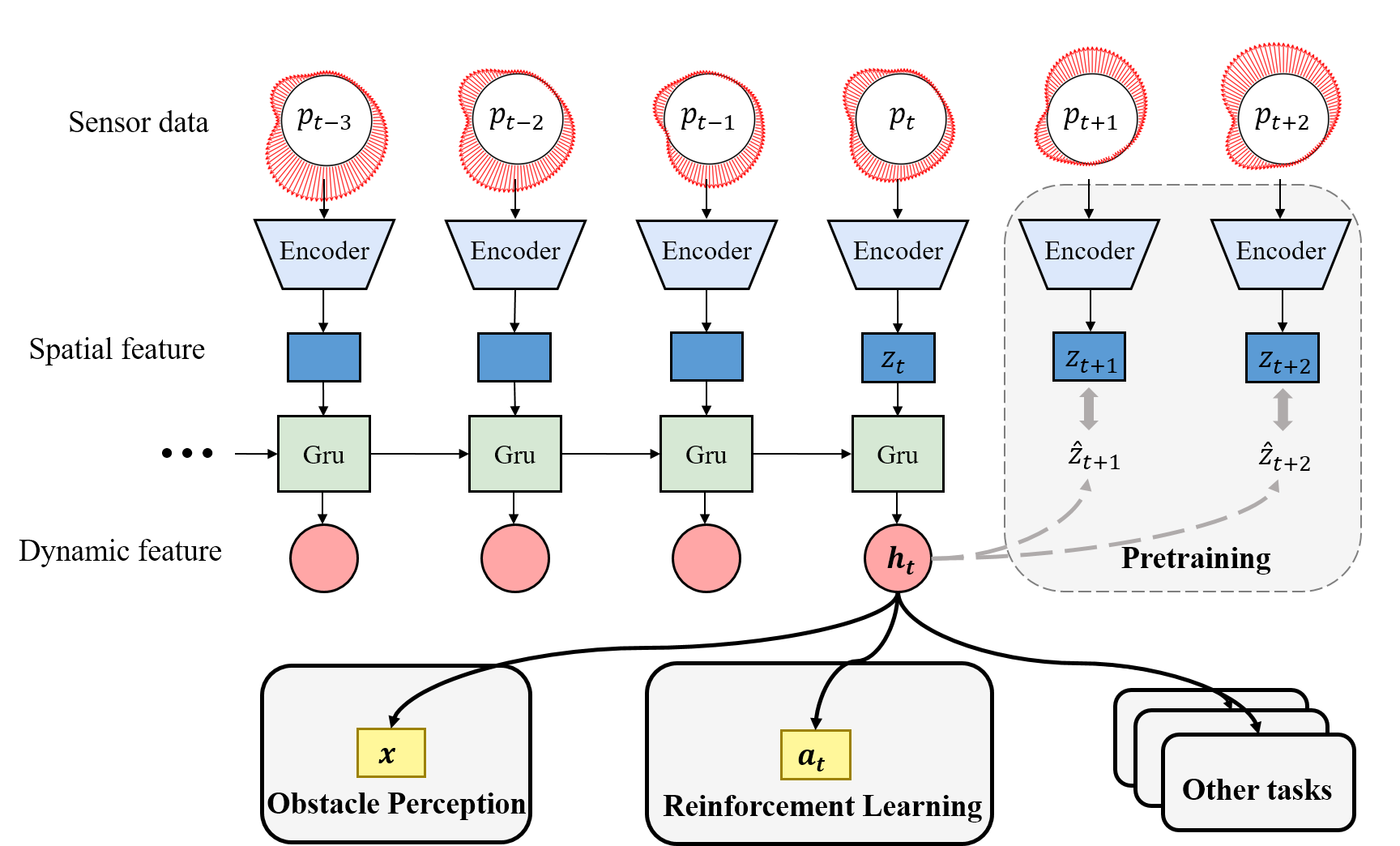}
\caption{\label{fig:3} Overview of the perceptual network architecture.}
\end{figure*}

\begin{figure}
\includegraphics[width=0.45\textwidth]{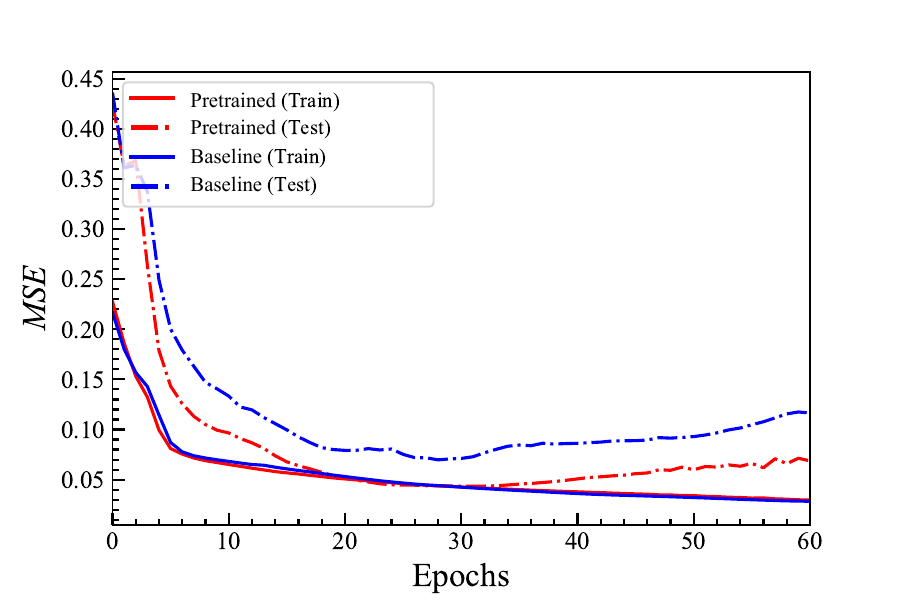}
\caption{\label{fig:training_loss} Loss curves for both pretrained and baseline agents.}
\end{figure}

\begin{figure*}
  \centering
  \subfloat[]{
    \includegraphics[width=0.45\textwidth]{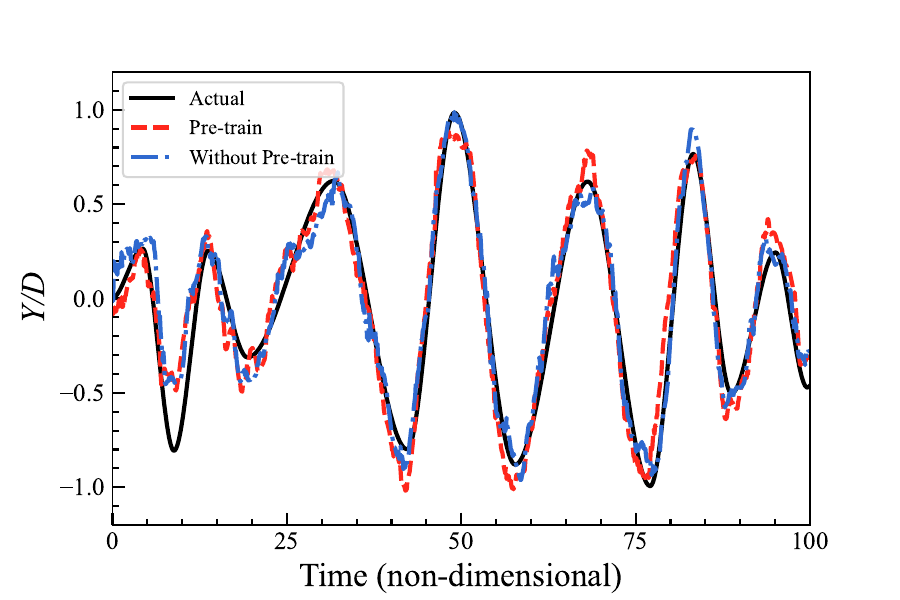}
    \label{fig4:subfig1}
  }
  \subfloat[]{
    \includegraphics[width=0.45\textwidth]{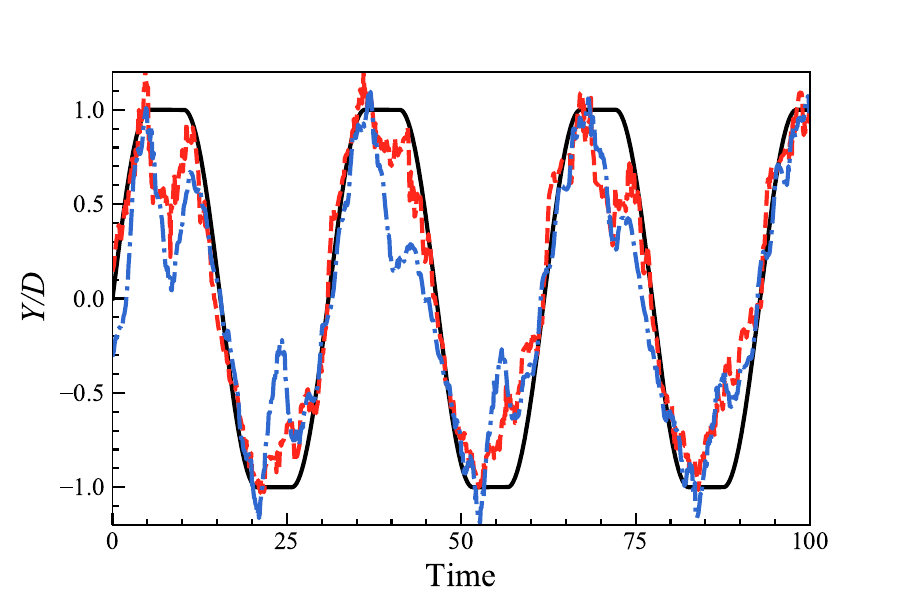}
    \label{fig4:subfig2}
  }
  \hfill
    \subfloat[]{
    \includegraphics[width=0.45\textwidth]{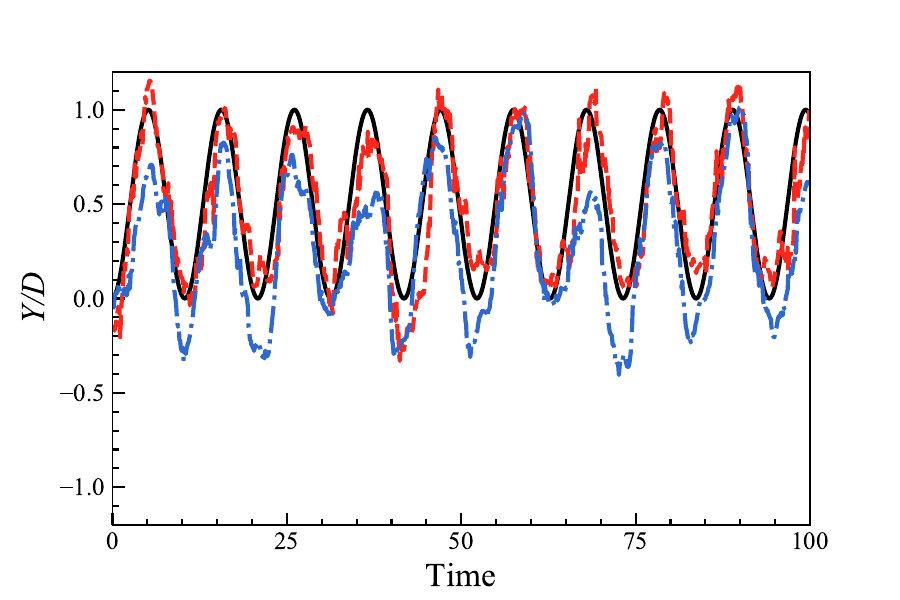}
    \label{fig4:subfig3}
  }
      \subfloat[]{
    \includegraphics[width=0.45\textwidth]{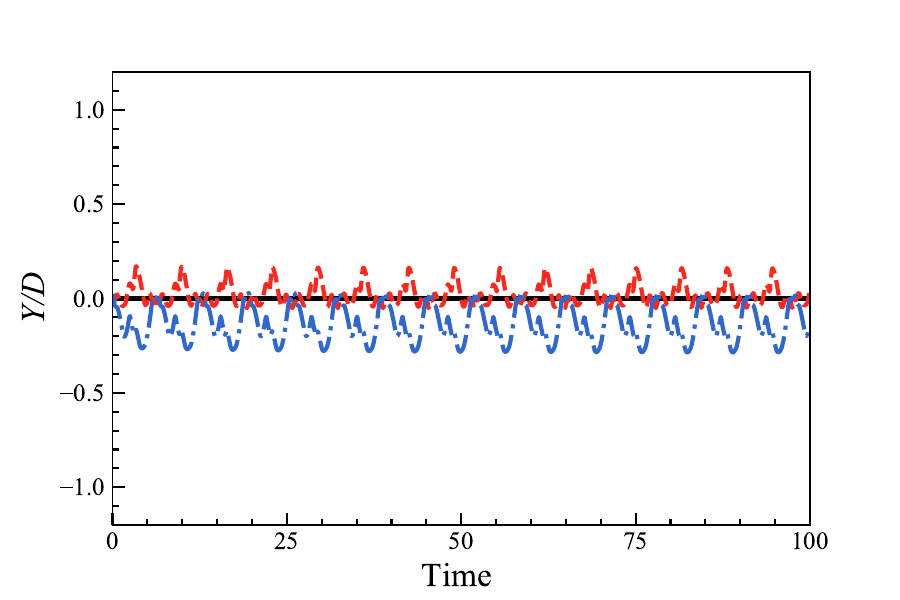}
    \label{fig4:subfig4}
  }
  \caption{Inferred versus real trajectories of the obstacle cylinder. (a) Test 1; (b) Test 2; (c) Test 3; (d) Test 4. }
  \label{fig:4}
\end{figure*}

\begin{figure}
\includegraphics[width=0.45\textwidth]{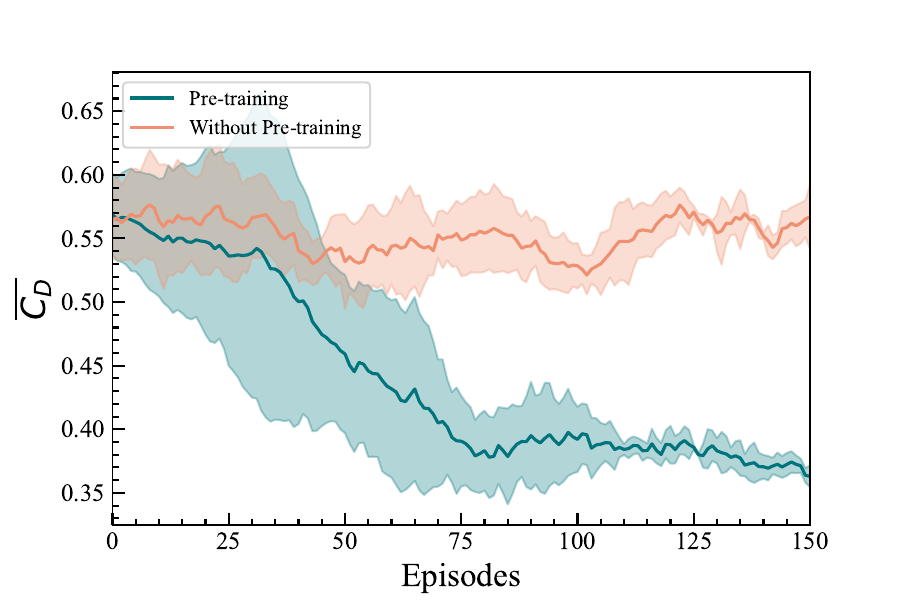}
\caption{\label{fig:learning_curves}Learning curves for pretrained and baseline agents. }
\end{figure}

\begin{figure*}
  \centering
  \subfloat[]{
    \includegraphics[width=0.4\textwidth]{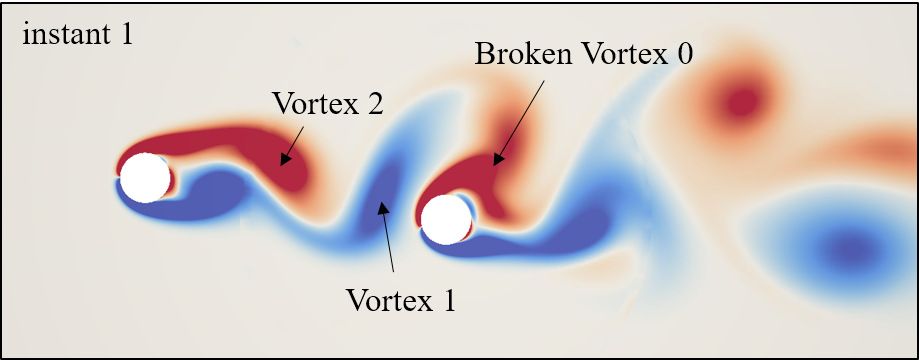}
    \label{fig5:subfig1}
  }
  \hspace{0.05\textwidth} 
  \subfloat[]{
    \includegraphics[width=0.155\textwidth]{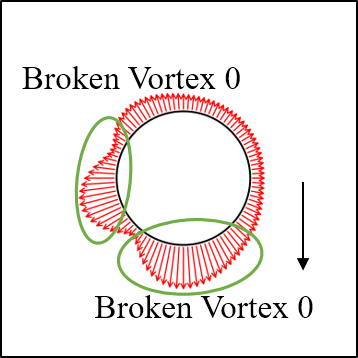}
    \label{fig5:subfig3}
  }

  \vspace{0.1in} 

  \subfloat[]{
    \includegraphics[width=0.4\textwidth]{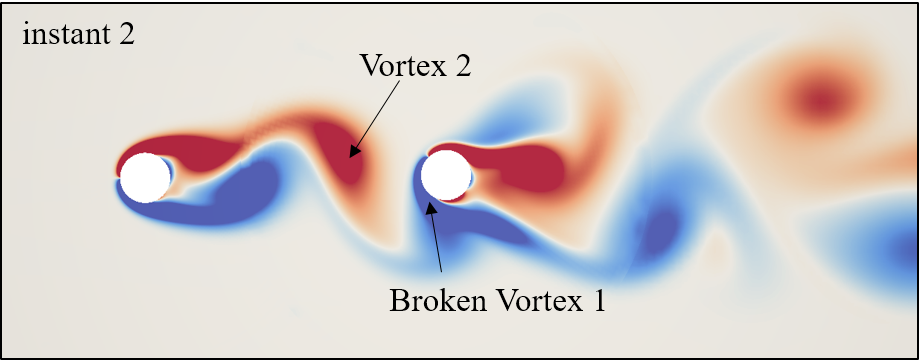}
    \label{fig5:subfig2}
  }
  \hspace{0.05\textwidth}
  \subfloat[]{
    \includegraphics[width=0.155\textwidth]{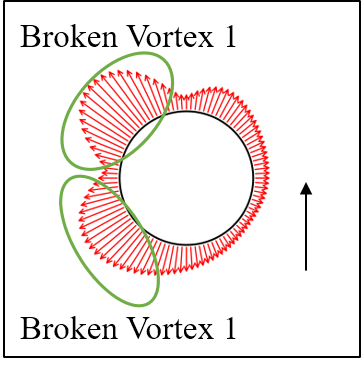}
    \label{fig5:subfig4}
  }

  \vspace{0.1in}

  \subfloat[]{
    \includegraphics[width=0.4\textwidth]{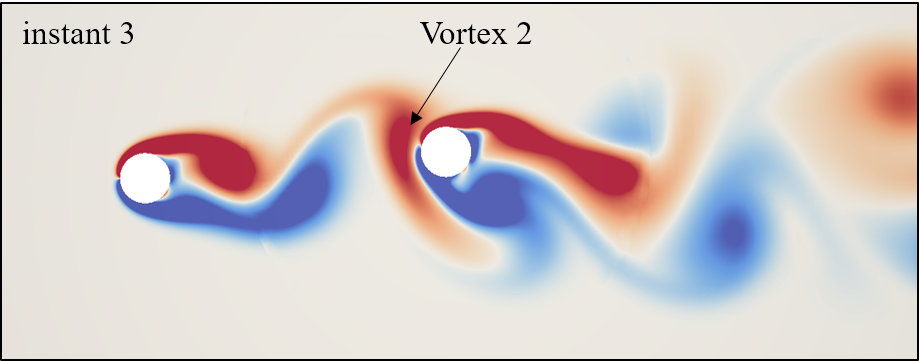}
    \label{fig5:subfig5}
  }
  \hspace{0.05\textwidth}
  \subfloat[]{
    \includegraphics[width=0.155\textwidth]{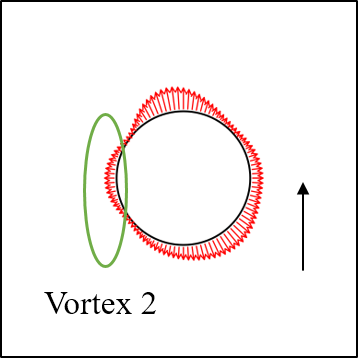}
    \label{fig5:subfig6}
  }

  \vspace{0.1in}

  \subfloat[]{
    \includegraphics[width=0.4\textwidth]{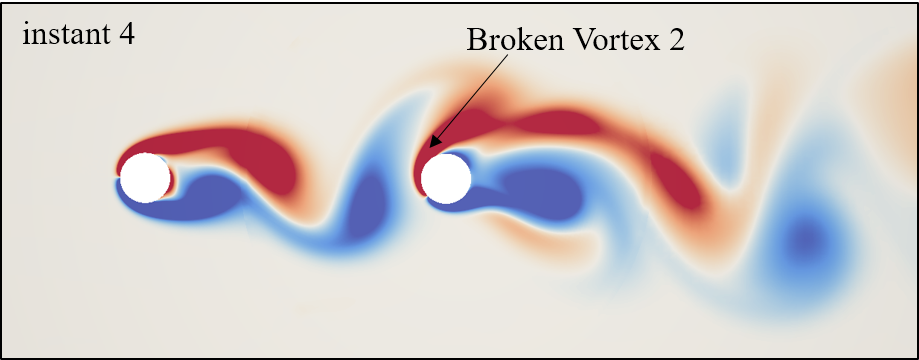}
    \label{fig5:subfig7}
  }
  \hspace{0.05\textwidth}
  \subfloat[]{
    \includegraphics[width=0.155\textwidth]{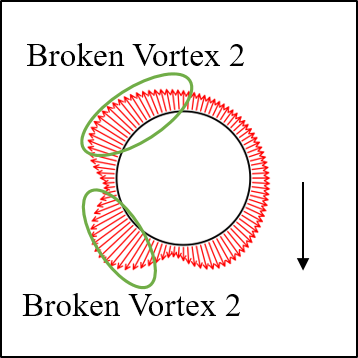}
    \label{fig5:subfig8}
  }

  \caption{Vorticity contours and surface pressure distribution at four time instants in the early RL training stage.}
  \label{fig:break_vortex}
\end{figure*}

\begin{figure*}
  \centering
  \subfloat[]{
    \includegraphics[width=0.4\textwidth]{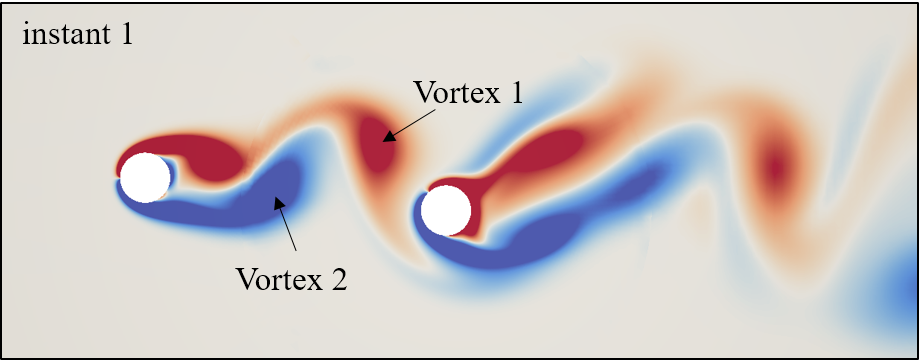}
    \label{fig6:subfig1}
  }
  \hspace{0.05\textwidth} 
  \subfloat[]{
    \includegraphics[width=0.155\textwidth]{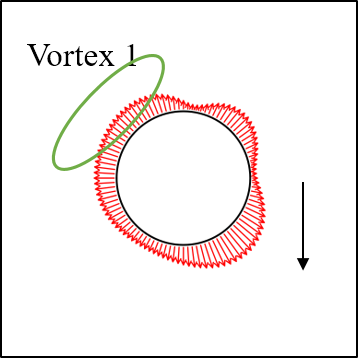}
    \label{fig6:subfig3}
  }

  \vspace{0.1in} 
  
  \subfloat[]{
    \includegraphics[width=0.4\textwidth]{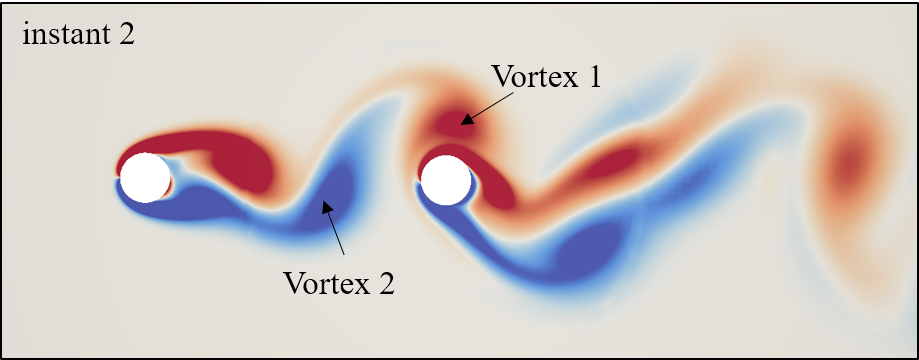}
    \label{fig6:subfig2}
  }
  \hspace{0.05\textwidth}
  \subfloat[]{
    \includegraphics[width=0.155\textwidth]{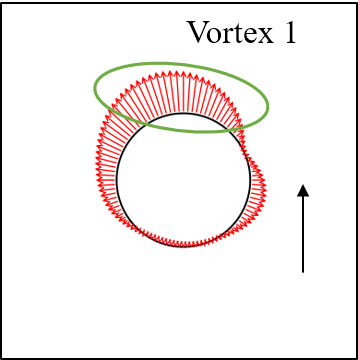}
    \label{fig6:subfig4}
  }
  
  \vspace{0.1in}
  \subfloat[]{
    \includegraphics[width=0.4\textwidth]{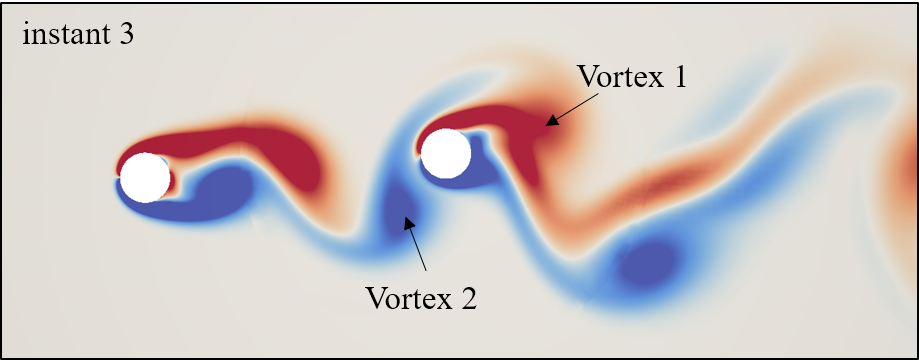}
    \label{fig6:subfig5}
  }
  \hspace{0.05\textwidth}
  \subfloat[]{
    \includegraphics[width=0.155\textwidth]{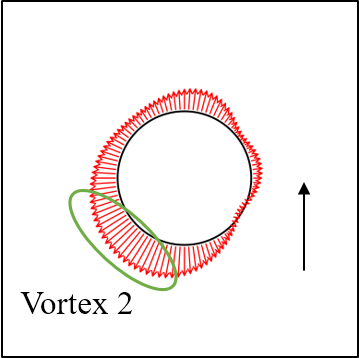}
    \label{fig6:subfig6}
  }
  \vspace{0.1in}
  
  \subfloat[]{
    \includegraphics[width=0.4\textwidth]{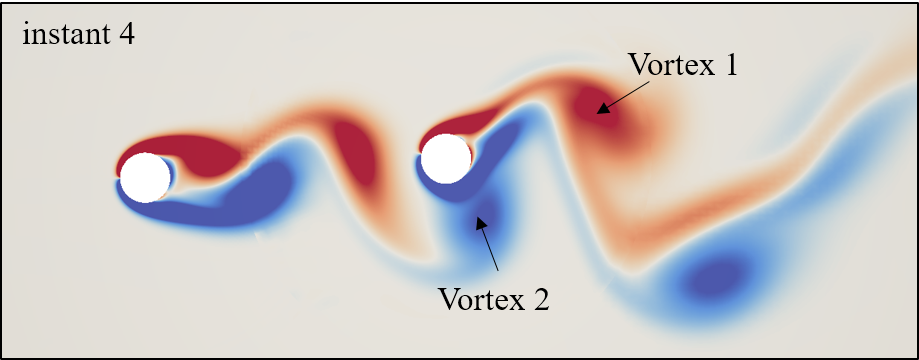}
    \label{fig6:subfig7}
  }
  \hspace{0.05\textwidth}
  \subfloat[]{
    \includegraphics[width=0.155\textwidth]{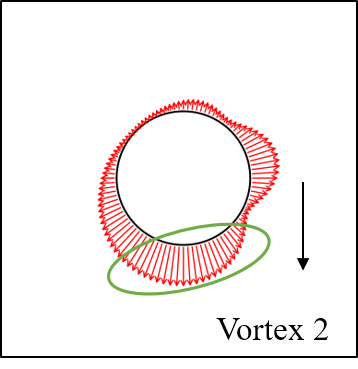}
    \label{fig6:subfig8}
  }
  
  \caption{Vorticity contours and surface pressure distribution at four time instants in the later RL training stages.}
  \label{fig:escape_vortex}
\end{figure*}

\begin{figure*}
  \centering
  \subfloat[]{
    \includegraphics[width=0.25\textwidth]{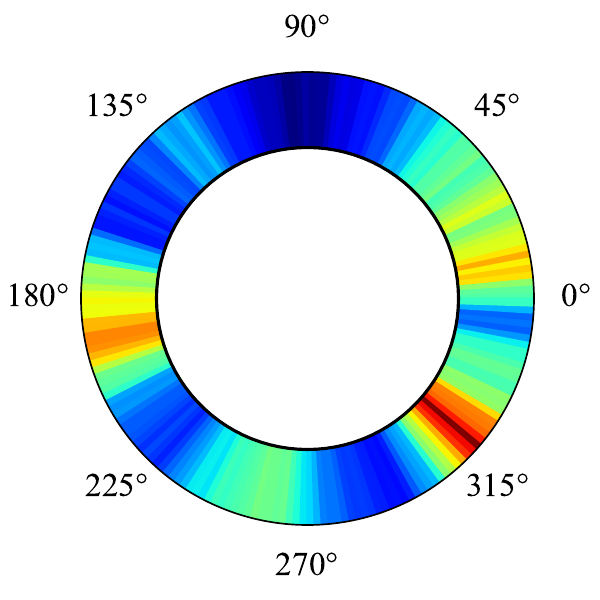}
    \label{fig8:subfig-1}
  }
  \subfloat[]{
    \includegraphics[width=0.25\textwidth]{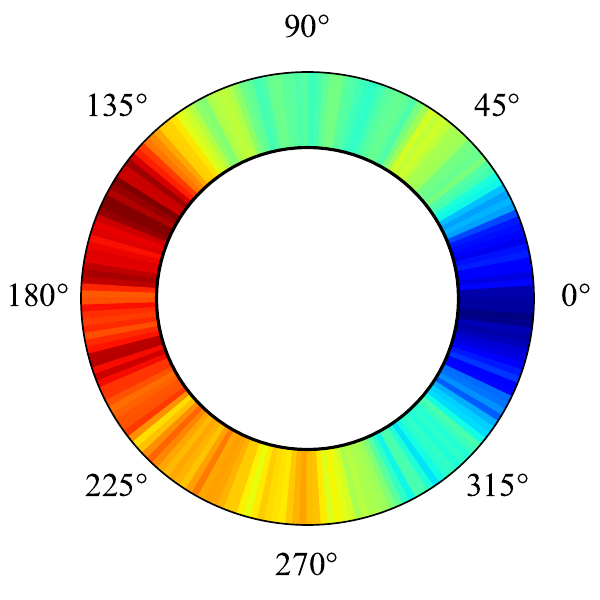}
    \label{fig8:subfig-2}
  }
    \subfloat[]{
    \includegraphics[width=0.25\textwidth]{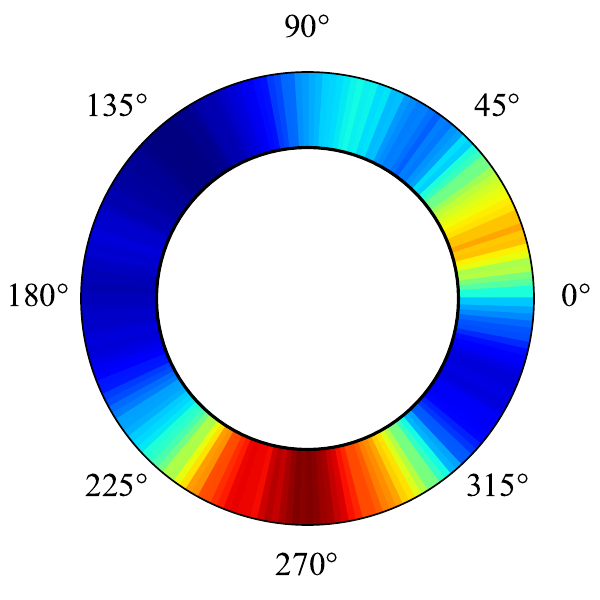}
    \label{fig8:subfig-3}
  }
  \hfill
  \subfloat[]{
    \includegraphics[width=0.25\textwidth]{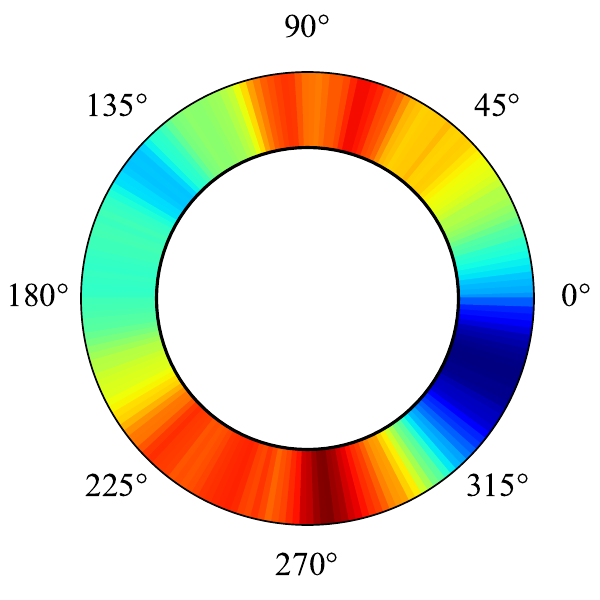}
    \label{fig8:subfig-4}
  }
    \subfloat[]{
    \includegraphics[width=0.25\textwidth]{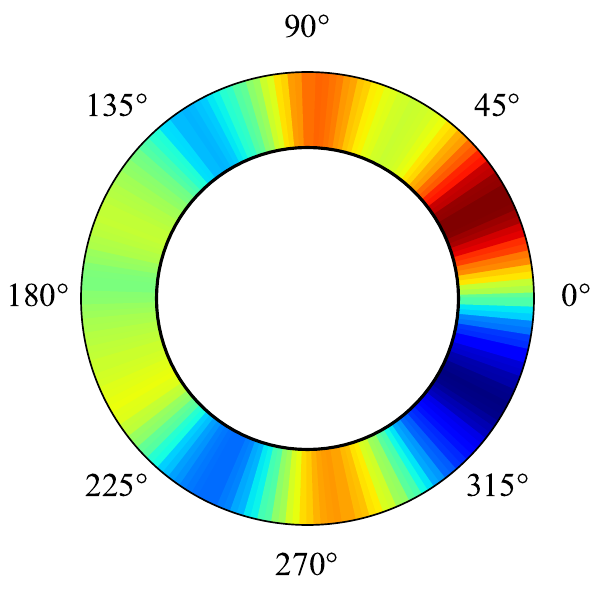}
    \label{fig8:subfig-5}
  }
    \subfloat[]{
    \includegraphics[width=0.25\textwidth]{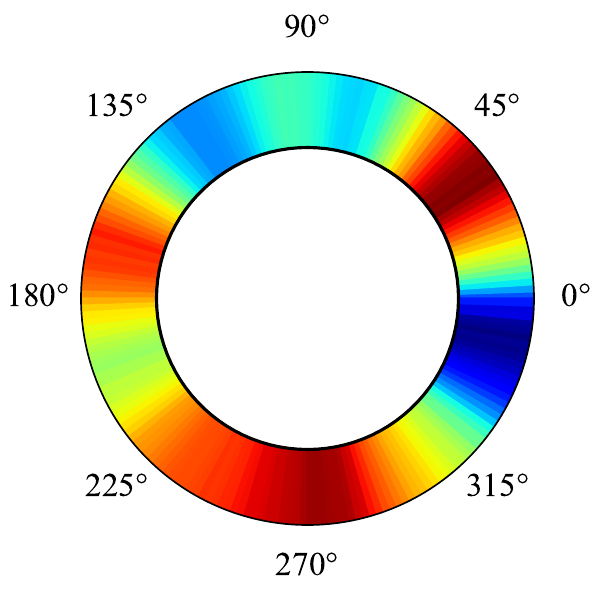}
    \label{fig8:subfig-6}
  }

  \caption{Sensitivity heatmaps of the agent cylinder surface. (a) Baseline agent; (b) Pretrained agent; (c) Baseline agent in obstacle perception task; (d) Pretrained agent in obstacle perception task; (e) Baseline agent in reinforcement learning task; (f) Pretrained agent in reinforcement learning task. Red areas indicate high sensitivity to pressure changes, while blue areas are less influential.}
  \label{fig:Gradient_map}
\end{figure*}

\subsection{Perceptual network}
The perceptual network consists of a convolutional neural network encoder and a GRU network \cite[]{ChoKyunghyun2014LPRu}. As shown in figure \ref{fig:3}, firstly, instant pressure data $p$ is compressed by convolutional neural network encoder, and instantaneous spatial feature $z$ is extracted. The spatial feature is then further compressed in time dimension by a GRU network, yielding the dynamic feature $h_t$ at time instant $t$. 

In this study, $p$ is a $1\times200$ vector sampled at 10 Hz. The spatial encoder consists of five convolutional layers with kernel sizes of [10, 8, 7, 5, 3] and strides of [2, 2, 1, 1, 1], and ReLU is used as the activation function. Then GRU network compresses spatial feature time series with size of $w\times50$ into a $1\times64$ dynamic feature vector $h$, where $w$ is the length of perceptual time window. 

During pretraining, a network with two linear layers maps $h_t$ to $\hat{z}_{t+1}$ and $\hat{z}_{t+2}$. The network is trained by maximizing the cosine similarity between $z_{t+1}$ and $\hat{z}_{t+1}$, and $z_{t+2}$ and $\hat{z}_{t+2}$, as defined below,

\begin{equation}
    \mathcal{L} = - \mathbb{E}_t \left[ \frac{\mathbf{z_{t+1}} \cdot \mathbf{\hat{z}_{t+1}}}{\left\lVert \mathbf{z_{t+1}} \right\rVert \left\lVert \mathbf{\hat{z}_{t+1}} \right\rVert} + 
    \frac{\mathbf{z_{t+2}} \cdot \mathbf{\hat{z}_{t+2}}}{\left\lVert \mathbf{z_{t+2}} \right\rVert \left\lVert \mathbf{\hat{z}_{t+2}} \right\rVert} \right]
\end{equation}
This process is similar to contrastive predictive coding \cite[]{vandenOordAaron2019RLwC}, which enables the agent to learn valuable knowledge by trying to forecast future information.

After pretraining, we anticipate the knowledge about surrounding fluid environment is well represented by the feature vector $h$ which can then be used in other tasks. For example, we can simply connect a network with linear layers to this feature vector to make predictions of flow quantities. Or we can use this feature vector as a state variable of given reinforcement learning model to optimize the performance of an intelligent agent.

\subsection{Reinforcement learning}\label{sec:Reinforcement learning}
A reinforcement learning problem is typically modeled as a Markov Decision Process (MDP). In an MDP, an agent interacts with an environment by taking actions and receiving feedback in the form of rewards. At each time step \( t \), the agent selects an action \( a_t \sim \pi_x (a|s_t) \) from a policy \( \pi \). This policy is parameterized by \( x \). Given an initial state \( s_0 \in S \), the agent takes an action and receives a reward \( r_{t+1} \). The objective is to learn a policy that maximizes the expected cumulative reward \( \mathbb{E}[\sum_{t=0}^{\infty} \gamma^t r_t|s_0,\pi_x] \), where \( \gamma \in [0,1) \) is a discount factor that prioritizes earlier rewards over later ones.

In this study, we employ the Proximal Policy Optimization (PPO) algorithm, developed by Schulman et al. \cite{SchulmanJohn2017PPOA}. PPO is an episodic learning method, which means learning process is divided into episodes. Each episode consists of a fixed number (set as 10 in current work) of actions. A key feature of PPO is to use Generalized Advantage Estimation (GAE) to reduce variance in the advantage estimates, which stabilizes training process. Additionally, PPO employs a clipping strategy to constrain policy updates, which ensures that the new policy does not deviate excessively from the old one, and thus a balance between exploration and exploitation is maintained. The PPO algorithm uses two neural networks: a policy network which decides the next action, and a value network that estimates the expected cumulative reward. Both networks include perceptual layers to process input data, followed by a fully connected layer to generate action or estimated advantage. In the training process, the learning rates for the policy network and for the value network are set as 0.0001 and 0.001, respectively.

\section{Results and discussion}
\subsection{Pretraining}
In the unsupervised pretraining process, the upstream obstacle cylinder oscillates randomly along the $y$-direction in the range of $-1D$ to $1D$ for 5000 seconds, while surface pressure of the downstream agent cylinder is sampled with a frequency of 10Hz. During pretraining, the perception network extracts dynamic features from past pressure data and uses them to predict spatial features $z$ for the next two time steps, as illustrated in figure \ref{fig:3}.

\subsection{Obstacle Perception}

Obstacle perception is a common task that could be encountered by embodied agents navigating in fluid environments. Disturbances originating from upstream obstacles propagate downstream, which can be perceived by downstream agents and used to estimate obstacle positions. In the current study, we design a supervised learning task to investigate obstacle perception ability of a pretrained agent. In this supervised learning task, the downstream agent is trained to reconstruct the disturbance trajectory of the upstream cylinder using surface pressure information of the agent cylinder. We build a training set by annotating the positions of the upstream cylinder over the 5000-second pretraining data set. Considering the time-delay of disturbance propagation, we use a 10-second window of pressure information to estimate the position of the obstacle cylinder 10 seconds ago.

For comparison, we carry out the same training process on agents both with and without pretraining. The network structures for both agents are identical, but the weights of the non-pretrained agent are randomly initialized. The non-pretrained model serves as the baseline model. After the obstacle perception training, we evaluate performance of both agents on four different test data sets.  In test set 1, the obstacle cylinder oscillates in the same random manner as in the training set, but the data sets are different. Test set 2 features intermittent motion with a 5-second pause at the end of each oscillation. Test set 3 involves sinusoidal motion restricted at one side of the flow region. In test set 4, the upstream obstacle cylinder is kept still at the centerline of the flow field.

In the training process, the errors for both training and test sets are calculated at each training step. The test error is calculated as the average of all 4 tests. Loss curves for both agents (pretrained and baseline) are plotted in figure \ref{fig:training_loss}. Figure \ref{fig:training_loss} shows that both agents perform similarly on the training set, while their performance on the test set differs significantly. The pretrained model achieves a minimum test set error of 0.0432, while the minimum error for the baseline model is 0.0696, which is 70$\%$ higher. Comparisons between perceived and real trajectories of the obstacle cylinder are shown in figure \ref{fig:4}. Both agents perform well on test set 1 (figure \ref{fig4:subfig1}), since the obstacle cylinder oscillates in the same random manner in this test as it does in the training process. However, on test sets 2, 3, and 4 (figures \ref{fig4:subfig2}, \ref{fig4:subfig3}, and \ref{fig4:subfig4}), the pretrained agent performs much better in perceiving the trajectory of obstacle cylinder. This demonstrates that the agent exhibits enhanced generalization capabilities in new scenarios after the pretraining process.

\subsection{Reinforcement Learning for drag reduction}
In the reinforcement learning task, the upstream obstacle cylinder is kept still. A Kármán vortex street is generated from the cylinder surface and sheds downstream with a Strouhal number of $St=fD/U=0.167$.  
Meanwhile, the agent cylinder is allowed to move in the $y$-direction in the range of -1D to 1D, and it keeps optimizing its motion trajectory in the vortical wake flow to minimize the drag exerted on it.

In the RL training process, the agent identifies instant dynamic feature of pressure time series data over past 10-second window and takes it as its state. This window length ensures sufficient information included in the state variable. The agent's action variable has two components: the next position of the cylinder $a_{pos}$ and the average velocity of the motion $a_{vel}$. We enforce the agent cylinder to cross the centerline of the flow field during each action. The reward function is defined as the negative value of the average drag induced by the current action. In this study, $a_{pos}$ ranges from -1 to 1, and $a_{vel}$ ranges from 0.2 to 0.4. In our setting, actions are sampled from a normal distribution, with the variance decreasing independently as the episodes progress.

Figure \ref{fig:learning_curves} presents the learning curves of the pretrained and baseline agents. The solid lines represent the mean values of five repeated RL training processes, while the shaded areas correspond to one standard deviation. As shown in figure \ref{fig:learning_curves}, initially the average drag coefficient of the agent cylinder with random motion is about 0.57. After 100 episodes of training, the pretrained agent significantly reduces its drag coefficient to a value below 0.37. In contrast, the baseline agent is trapped in the complexity of the problem and fails to find a motion pattern to reduce the drag acting on it.  

Performance evolution of pretrained agent can be better illustrated by flow field snapshots taken in the RL training process. Figure \ref{fig:break_vortex} shows flow field snapshots captured at 4 continuous time instants in the early stage of RL training process on the pretrained agent. The time interval between snapshots is 2 seconds. As shown in figure \ref{fig:break_vortex}, in the early stage of training, the agent cylinder oscillates aimlessly in $y$-direction and it keeps colliding with the shedding vortices from upstream. The breakdown of shedding vortices induces additional drag force on the agent cylinder. However, through tens of episodes of trial-and-error, the agent is able to adjust its action strategy using pretrained knowledge. As depicted in flow field snapshots obtained from the later stage of training process (figure \ref{fig:escape_vortex}), the agent cylinder chooses to move through the gaps between vortices to avoid direct collision with shedding vortices, which significantly reduces the drag force exerted on the cylinder.

\subsection{Sensitivity Analysis}
Previous sections have shown that the agent with pretraining outperforms the baseline agent in both obstacle perception and reinforcement learning tasks. To further analyze this performance difference, we compute the sensitivity of the dynamic feature $h_t$ with respect to each surface pressure input and plot the distribution of sensitivity values over the agent cylinder surface in figure \ref{fig:Gradient_map}. This sensitivity distribution reveals which areas of the surface pressure information on the agent cylinder the perception network focuses on. Before pretraining, with initially randomized parameters, the perception network's understanding of the surrounding fluid environment is chaotic and lacks a focused region (figure \ref{fig8:subfig-1}). After pretraining, the perception network adjusts its concentration to the upstream side of the cylinder which indicates this area is important in the pretraining task---the prediction of future spatial features of surface pressure distribution  (figure \ref{fig8:subfig-2}). 

In the following specific tasks, the agent further adjusts parameters of the perception network to fit objectives of the tasks. For example, in the obstacle perception task (figure \ref{fig8:subfig-3}, \ref{fig8:subfig-4}), sensitivity concentration region is shifted from the upstream side of the cylinder to both lateral sides where the shedding vortices might hit the cylinder. In contrast, sensitivity distribution of the baseline agent only focuses on one side of the cylinder, which might result in worse performance in the obstacle perception task. Furthermore, in the reinforcement learning task (figure \ref{fig8:subfig-5}, \ref{fig8:subfig-6}), the pretrained agent distributes its attention to upstream, downstream and lateral sides of the cylinder. In this complex task, the agent needs to perceive not only the incoming flow but also the pressure variance induced by the lateral motion of the cylinder. On the other hand, the baseline agent mainly focuses on the downstream and slightly on the lateral sides of the cylinder. A lack of information on lateral and upstream sides of the cylinder might lead to the failure of the baseline agent in this drag reduction task.

The sensitivity analysis above shows that the pretraining process enables the agent to adjust the parameters of the perception network to fit new task scenarios better and faster, thus outperforming the agent with a randomly initialized network.

\section{Conclusion}
We develop a perception network that can be used for information compression in fluid environments. The network consists of a convolutional neural network for spatial compression and a GRU network for temporal compression. We propose a pretraining process which is to predict spatial feature variables of next two time instants. 

We have tested the perception network and the pretraining process in a two-cylinder problem. The results show that the pretrained agent cylinder can better perceive the position of the upstream obstacle cylinder. Moreover, in the subsequent reinforcement learning task, the baseline agent fails to develop an effective strategy for drag reduction. On the other hand, through trials and errors, the pretrained agent successfully finds a motion strategy to avoid shedding vortices in the Kármán vortex street, which results in significant drag reduction. The sensitivity analysis indicates that this performance could be due to the better attention distribution of the pretrained agent. 

This study demonstrates that the pretraining process with the perception network in fluid environments can significantly improve the agent's performance in subsequent tasks, offering a promising pathway toward a robust and effective solution for real-world, complex, multi-scenario fluid problems.

\section*{Data Availability Statement}
The data that support the findings of this study are avail-able from the corresponding author upon reasonablerequest.

\section*{Disclosure statement}
The authors report there are no competing interests to declare.

\appendix
\section{Validation of the fluid solver}\label{Appendix A}
\begin{figure}
  \centering
  \subfloat[]{
    \includegraphics[width=0.45\textwidth]{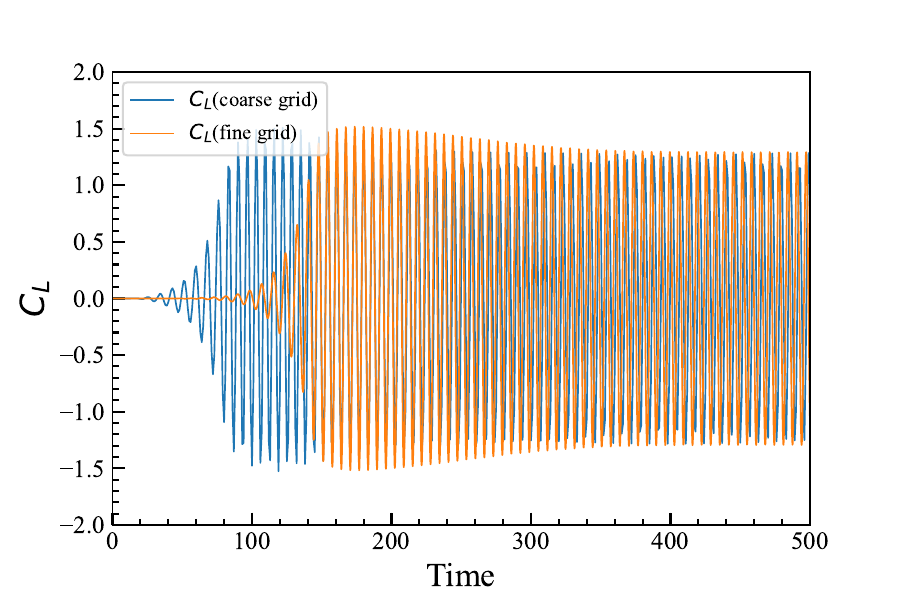}
    \label{fig10:subfig-a}
  }
  \hfill
  \subfloat[]{
    \includegraphics[width=0.45\textwidth]{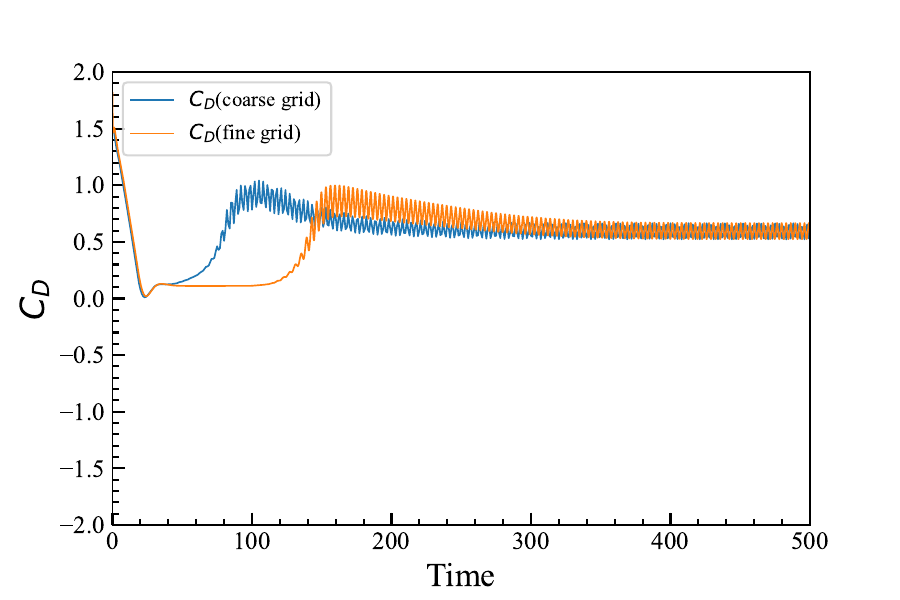}
    \label{fig10:subfig-b}
  }
\caption{\label{fig:mesh_converge}Mesh Convergence Test.}
\end{figure}

In numerical simulation, time step is chosen so that the maximum Courant number during computation does not exceed 0.5. A mesh convergence test is carried out to confirm the mesh size. In mesh convergence test, flow field around two fixed cylinders is computed and both both coarse (67,352 elements) and fine grids (89,400 elements) are used. The time histories of drag and lift coefficients of the downstream agent cylinder are recorded and plotted in figure \ref{fig:mesh_converge}. The results demonstrate spatial convergence on the fine grid, which is used in subsequent simulations.

We validate the fluid solver by computing flow field around single cylinder. The computed average drag and maximum lift coefficients at $Re=100$ are compared with previous literature, as shown in Table \ref{tab:table1}. The comparison of variance of Strouhal number versus Reynolds number is shown in Table \ref{tab:table2}. 

To further validate the accuracy of the overset mesh technique in simulating flow around moving cylinder, additional simulations are performed on an in-line oscillating cylinder in uniform flow at a Reynolds number of 100. The cylinder oscillates parallel to the free stream at a frequency twice of the vortex shedding frequency for a fixed cylinder, with an oscillation amplitude of 0.14 times the cylinder diameter $D$. Table \ref{tab:table3} shows that the present results are consistent with those reported in the earlier literature.

\begin{table*}
\caption{\label{tab:table1}The comparison of maximum lift coefficient and mean drag coefcient at Re = 100}
\begin{ruledtabular}
\begin{tabular}{lccccc}
 & Present  &Shu et al.\cite{ShuC.2007Anib}& Liu et al.\cite{LIU199835} & Su et al.\cite{SuShen-Wei2007Aibt} &Tseng and Ferziger \cite{TsengYu-Heng2003Agib}  \\
\hline
$C_L$ & 0.344 & 0.350 & 0.339&  0.340 &0.290\\
$C_D$ & 1.342  & 1.383 & 1.350& 1.400 &1.420\\
\end{tabular}
\end{ruledtabular}
\end{table*}

\begin{table}
\caption{\label{tab:table2}The comparison of Strouhal number for different Re}
\begin{ruledtabular}
\begin{tabular}{lccc}
Re&Present& Su et al.\cite{SuShen-Wei2007Aibt} &Williamson\cite{WilliamsonC.H.K.1988Daua} \\
\hline
80 & 0.153 & 0.153 & 0.15\\
100 & 0.167 & 0.168 &0.166\\
150 & 0.183 & 0.187 &0.183\\
\end{tabular}
\end{ruledtabular}
\end{table}

\begin{table}
\caption{\label{tab:table3}The comparisons of lift and drag coefficients of in-line oscillating cylinder}
\begin{ruledtabular}
\begin{tabular}{lccc}
 &Present& Su et al.\cite{SuShen-Wei2007Aibt} &Hurlbut et al.\cite{HurlbutS.E1982NSfL} \\
\hline
$C_L$ & 0.91 & 0.97 & 0.95\\
$C_D$ & 1.71 & 1.70 & 1.68\\
\end{tabular}
\end{ruledtabular}
\end{table}

\bibliography{aipsamp}

\begin{thebibliography}{33}%
\makeatletter
\providecommand \@ifxundefined [1]{%
 \@ifx{#1\undefined}
}%
\providecommand \@ifnum [1]{%
 \ifnum #1\expandafter \@firstoftwo
 \else \expandafter \@secondoftwo
 \fi
}%
\providecommand \@ifx [1]{%
 \ifx #1\expandafter \@firstoftwo
 \else \expandafter \@secondoftwo
 \fi
}%
\providecommand \natexlab [1]{#1}%
\providecommand \enquote  [1]{``#1''}%
\providecommand \bibnamefont  [1]{#1}%
\providecommand \bibfnamefont [1]{#1}%
\providecommand \citenamefont [1]{#1}%
\providecommand \href@noop [0]{\@secondoftwo}%
\providecommand \href [0]{\begingroup \@sanitize@url \@href}%
\providecommand \@href[1]{\@@startlink{#1}\@@href}%
\providecommand \@@href[1]{\endgroup#1\@@endlink}%
\providecommand \@sanitize@url [0]{\catcode `\\12\catcode `\$12\catcode `\&12\catcode `\#12\catcode `\^12\catcode `\_12\catcode `\%12\relax}%
\providecommand \@@startlink[1]{}%
\providecommand \@@endlink[0]{}%
\providecommand \url  [0]{\begingroup\@sanitize@url \@url }%
\providecommand \@url [1]{\endgroup\@href {#1}{\urlprefix }}%
\providecommand \urlprefix  [0]{URL }%
\providecommand \Eprint [0]{\href }%
\providecommand \doibase [0]{http://dx.doi.org/}%
\providecommand \selectlanguage [0]{\@gobble}%
\providecommand \bibinfo  [0]{\@secondoftwo}%
\providecommand \bibfield  [0]{\@secondoftwo}%
\providecommand \translation [1]{[#1]}%
\providecommand \BibitemOpen [0]{}%
\providecommand \bibitemStop [0]{}%
\providecommand \bibitemNoStop [0]{.\EOS\space}%
\providecommand \EOS [0]{\spacefactor3000\relax}%
\providecommand \BibitemShut  [1]{\csname bibitem#1\endcsname}%
\let\auto@bib@innerbib\@empty
\bibitem [{\citenamefont {Smith}\ and\ \citenamefont {Gasser}(2005)}]{SmithLinda2005TDoE}%
  \BibitemOpen
  \bibfield  {author} {\bibinfo {author} {\bibfnamefont {L.}~\bibnamefont {Smith}}\ and\ \bibinfo {author} {\bibfnamefont {M.}~\bibnamefont {Gasser}},\ }\bibfield  {title} {\enquote {\bibinfo {title} {The development of embodied cognition: Six lessons from babies},}\ }\href@noop {} {\bibfield  {journal} {\bibinfo  {journal} {Artificial life}\ }\textbf {\bibinfo {volume} {11}},\ \bibinfo {pages} {13--29} (\bibinfo {year} {2005})}\BibitemShut {NoStop}%
\bibitem [{\citenamefont {Duan}\ \emph {et~al.}(2022)\citenamefont {Duan}, \citenamefont {Yu}, \citenamefont {Tan}, \citenamefont {Zhu},\ and\ \citenamefont {Tan}}]{DuanJiafei2022ASoE}%
  \BibitemOpen
  \bibfield  {author} {\bibinfo {author} {\bibfnamefont {J.}~\bibnamefont {Duan}}, \bibinfo {author} {\bibfnamefont {S.}~\bibnamefont {Yu}}, \bibinfo {author} {\bibfnamefont {H.~L.}\ \bibnamefont {Tan}}, \bibinfo {author} {\bibfnamefont {H.}~\bibnamefont {Zhu}}, \ and\ \bibinfo {author} {\bibfnamefont {C.}~\bibnamefont {Tan}},\ }\bibfield  {title} {\enquote {\bibinfo {title} {A survey of embodied ai: From simulators to research tasks},}\ }\href@noop {} {\bibfield  {journal} {\bibinfo  {journal} {IEEE transactions on emerging topics in computational intelligence}\ }\textbf {\bibinfo {volume} {6}},\ \bibinfo {pages} {230--244} (\bibinfo {year} {2022})}\BibitemShut {NoStop}%
\bibitem [{\citenamefont {Brunton}, \citenamefont {Noack},\ and\ \citenamefont {Koumoutsakos}(2020)}]{BruntonStevenL2020MLfF}%
  \BibitemOpen
  \bibfield  {author} {\bibinfo {author} {\bibfnamefont {S.~L.}\ \bibnamefont {Brunton}}, \bibinfo {author} {\bibfnamefont {B.~R.}\ \bibnamefont {Noack}}, \ and\ \bibinfo {author} {\bibfnamefont {P.}~\bibnamefont {Koumoutsakos}},\ }\bibfield  {title} {\enquote {\bibinfo {title} {Machine learning for fluid mechanics},}\ }\href@noop {} {\bibfield  {journal} {\bibinfo  {journal} {Annual review of fluid mechanics}\ }\textbf {\bibinfo {volume} {52}},\ \bibinfo {pages} {477--508} (\bibinfo {year} {2020})}\BibitemShut {NoStop}%
\bibitem [{\citenamefont {Vignon}, \citenamefont {Rabault},\ and\ \citenamefont {Vinuesa}(2023)}]{VignonC2023Raia}%
  \BibitemOpen
  \bibfield  {author} {\bibinfo {author} {\bibfnamefont {C.}~\bibnamefont {Vignon}}, \bibinfo {author} {\bibfnamefont {J.}~\bibnamefont {Rabault}}, \ and\ \bibinfo {author} {\bibfnamefont {R.}~\bibnamefont {Vinuesa}},\ }\bibfield  {title} {\enquote {\bibinfo {title} {Recent advances in applying deep reinforcement learning for flow control: Perspectives and future directions},}\ }\href@noop {} {\bibfield  {journal} {\bibinfo  {journal} {Physics of Fluids}\ }\textbf {\bibinfo {volume} {35}} (\bibinfo {year} {2023})}\BibitemShut {NoStop}%
\bibitem [{\citenamefont {Reddy}\ \emph {et~al.}(2018)\citenamefont {Reddy}, \citenamefont {Wong-Ng}, \citenamefont {Celani}, \citenamefont {Sejnowski},\ and\ \citenamefont {Vergassola}}]{ReddyGautam2018Gsvr}%
  \BibitemOpen
  \bibfield  {author} {\bibinfo {author} {\bibfnamefont {G.}~\bibnamefont {Reddy}}, \bibinfo {author} {\bibfnamefont {J.}~\bibnamefont {Wong-Ng}}, \bibinfo {author} {\bibfnamefont {A.}~\bibnamefont {Celani}}, \bibinfo {author} {\bibfnamefont {T.~J.}\ \bibnamefont {Sejnowski}}, \ and\ \bibinfo {author} {\bibfnamefont {M.}~\bibnamefont {Vergassola}},\ }\bibfield  {title} {\enquote {\bibinfo {title} {Glider soaring via reinforcement learning in the field},}\ }\href@noop {} {\bibfield  {journal} {\bibinfo  {journal} {Nature}\ }\textbf {\bibinfo {volume} {562}},\ \bibinfo {pages} {236--239} (\bibinfo {year} {2018})}\BibitemShut {NoStop}%
\bibitem [{\citenamefont {Novati}, \citenamefont {Mahadevan},\ and\ \citenamefont {Koumoutsakos}(2019)}]{NovatiGuido2019Cgap}%
  \BibitemOpen
  \bibfield  {author} {\bibinfo {author} {\bibfnamefont {G.}~\bibnamefont {Novati}}, \bibinfo {author} {\bibfnamefont {L.}~\bibnamefont {Mahadevan}}, \ and\ \bibinfo {author} {\bibfnamefont {P.}~\bibnamefont {Koumoutsakos}},\ }\bibfield  {title} {\enquote {\bibinfo {title} {Controlled gliding and perching through deep-reinforcement-learning},}\ }\href@noop {} {\bibfield  {journal} {\bibinfo  {journal} {Physical review fluids}\ }\textbf {\bibinfo {volume} {4}} (\bibinfo {year} {2019})}\BibitemShut {NoStop}%
\bibitem [{\citenamefont {Rabault}\ \emph {et~al.}(2019)\citenamefont {Rabault}, \citenamefont {Kuchta}, \citenamefont {Jensen}, \citenamefont {Réglade},\ and\ \citenamefont {Cerardi}}]{RabaultJean2019Annt}%
  \BibitemOpen
  \bibfield  {author} {\bibinfo {author} {\bibfnamefont {J.}~\bibnamefont {Rabault}}, \bibinfo {author} {\bibfnamefont {M.}~\bibnamefont {Kuchta}}, \bibinfo {author} {\bibfnamefont {A.}~\bibnamefont {Jensen}}, \bibinfo {author} {\bibfnamefont {U.}~\bibnamefont {Réglade}}, \ and\ \bibinfo {author} {\bibfnamefont {N.}~\bibnamefont {Cerardi}},\ }\bibfield  {title} {\enquote {\bibinfo {title} {Artificial neural networks trained through deep reinforcement learning discover control strategies for active flow control},}\ }\href@noop {} {\bibfield  {journal} {\bibinfo  {journal} {Journal of fluid mechanics}\ }\textbf {\bibinfo {volume} {865}},\ \bibinfo {pages} {281--302} (\bibinfo {year} {2019})}\BibitemShut {NoStop}%
\bibitem [{\citenamefont {Fan}\ \emph {et~al.}(2020)\citenamefont {Fan}, \citenamefont {Yang}, \citenamefont {Wang}, \citenamefont {Triantafyllou},\ and\ \citenamefont {Karniadakis}}]{FanDixia2020Rlfb}%
  \BibitemOpen
  \bibfield  {author} {\bibinfo {author} {\bibfnamefont {D.}~\bibnamefont {Fan}}, \bibinfo {author} {\bibfnamefont {L.}~\bibnamefont {Yang}}, \bibinfo {author} {\bibfnamefont {Z.}~\bibnamefont {Wang}}, \bibinfo {author} {\bibfnamefont {M.~S.}\ \bibnamefont {Triantafyllou}}, \ and\ \bibinfo {author} {\bibfnamefont {G.~E.}\ \bibnamefont {Karniadakis}},\ }\bibfield  {title} {\enquote {\bibinfo {title} {Reinforcement learning for bluff body active flow control in experiments and simulations},}\ }\href@noop {} {\bibfield  {journal} {\bibinfo  {journal} {Proceedings of the National Academy of Sciences - PNAS}\ }\textbf {\bibinfo {volume} {117}},\ \bibinfo {pages} {26091--26098} (\bibinfo {year} {2020})}\BibitemShut {NoStop}%
\bibitem [{\citenamefont {Gazzola}, \citenamefont {Hejazialhosseini},\ and\ \citenamefont {Koumoutsakos}(2014)}]{GazzolaMattia2014RLaW}%
  \BibitemOpen
  \bibfield  {author} {\bibinfo {author} {\bibfnamefont {M.}~\bibnamefont {Gazzola}}, \bibinfo {author} {\bibfnamefont {B.}~\bibnamefont {Hejazialhosseini}}, \ and\ \bibinfo {author} {\bibfnamefont {P.}~\bibnamefont {Koumoutsakos}},\ }\bibfield  {title} {\enquote {\bibinfo {title} {Reinforcement learning and wavelet adapted vortex methods for simulations of self-propelled swimmers},}\ }\href@noop {} {\bibfield  {journal} {\bibinfo  {journal} {SIAM journal on scientific computing}\ }\textbf {\bibinfo {volume} {36}},\ \bibinfo {pages} {B622--B639} (\bibinfo {year} {2014})}\BibitemShut {NoStop}%
\bibitem [{\citenamefont {Verma}, \citenamefont {Novati},\ and\ \citenamefont {Koumoutsakos}(2018)}]{VermaSiddhartha2018Ecsb}%
  \BibitemOpen
  \bibfield  {author} {\bibinfo {author} {\bibfnamefont {S.}~\bibnamefont {Verma}}, \bibinfo {author} {\bibfnamefont {G.}~\bibnamefont {Novati}}, \ and\ \bibinfo {author} {\bibfnamefont {P.}~\bibnamefont {Koumoutsakos}},\ }\bibfield  {title} {\enquote {\bibinfo {title} {Efficient collective swimming by harnessing vortices through deep reinforcement learning},}\ }\href@noop {} {\bibfield  {journal} {\bibinfo  {journal} {Proceedings of the National Academy of Sciences - PNAS}\ }\textbf {\bibinfo {volume} {115}},\ \bibinfo {pages} {5849--5854} (\bibinfo {year} {2018})}\BibitemShut {NoStop}%
\bibitem [{\citenamefont {Zhu}\ \emph {et~al.}(2021)\citenamefont {Zhu}, \citenamefont {Tian}, \citenamefont {Young}, \citenamefont {Liao},\ and\ \citenamefont {Lai}}]{ZhuYi2021Anso}%
  \BibitemOpen
  \bibfield  {author} {\bibinfo {author} {\bibfnamefont {Y.}~\bibnamefont {Zhu}}, \bibinfo {author} {\bibfnamefont {F.-B.}\ \bibnamefont {Tian}}, \bibinfo {author} {\bibfnamefont {J.}~\bibnamefont {Young}}, \bibinfo {author} {\bibfnamefont {J.~C.}\ \bibnamefont {Liao}}, \ and\ \bibinfo {author} {\bibfnamefont {J.~C.~S.}\ \bibnamefont {Lai}},\ }\bibfield  {title} {\enquote {\bibinfo {title} {A numerical study of fish adaption behaviors in complex environments with a deep reinforcement learning and immersed boundary-lattice boltzmann method},}\ }\href@noop {} {\bibfield  {journal} {\bibinfo  {journal} {Scientific reports}\ }\textbf {\bibinfo {volume} {11}},\ \bibinfo {pages} {1691--1691} (\bibinfo {year} {2021})}\BibitemShut {NoStop}%
\bibitem [{\citenamefont {Paris}, \citenamefont {Beneddine},\ and\ \citenamefont {Dandois}(2021)}]{ParisRomain2021Rfca}%
  \BibitemOpen
  \bibfield  {author} {\bibinfo {author} {\bibfnamefont {R.}~\bibnamefont {Paris}}, \bibinfo {author} {\bibfnamefont {S.}~\bibnamefont {Beneddine}}, \ and\ \bibinfo {author} {\bibfnamefont {J.}~\bibnamefont {Dandois}},\ }\bibfield  {title} {\enquote {\bibinfo {title} {Robust flow control and optimal sensor placement using deep reinforcement learning},}\ }\href@noop {} {\bibfield  {journal} {\bibinfo  {journal} {Journal of fluid mechanics}\ }\textbf {\bibinfo {volume} {913}},\ \bibinfo {pages} {A25} (\bibinfo {year} {2021})}\BibitemShut {NoStop}%
\bibitem [{\citenamefont {Li}\ and\ \citenamefont {Zhang}(2022)}]{LiJichao2022Rcoc}%
  \BibitemOpen
  \bibfield  {author} {\bibinfo {author} {\bibfnamefont {J.}~\bibnamefont {Li}}\ and\ \bibinfo {author} {\bibfnamefont {M.}~\bibnamefont {Zhang}},\ }\bibfield  {title} {\enquote {\bibinfo {title} {Reinforcement-learning-based control of confined cylinder wakes with stability analyses},}\ }\href@noop {} {\bibfield  {journal} {\bibinfo  {journal} {Journal of fluid mechanics}\ }\textbf {\bibinfo {volume} {932}},\ \bibinfo {pages} {A44} (\bibinfo {year} {2022})}\BibitemShut {NoStop}%
\bibitem [{\citenamefont {Wang}\ \emph {et~al.}(2024)\citenamefont {Wang}, \citenamefont {Yan}, \citenamefont {Hu}, \citenamefont {Chen}, \citenamefont {Rabault},\ and\ \citenamefont {Noack}}]{WangQiulei2024Dfdr}%
  \BibitemOpen
  \bibfield  {author} {\bibinfo {author} {\bibfnamefont {Q.}~\bibnamefont {Wang}}, \bibinfo {author} {\bibfnamefont {L.}~\bibnamefont {Yan}}, \bibinfo {author} {\bibfnamefont {G.}~\bibnamefont {Hu}}, \bibinfo {author} {\bibfnamefont {W.}~\bibnamefont {Chen}}, \bibinfo {author} {\bibfnamefont {J.}~\bibnamefont {Rabault}}, \ and\ \bibinfo {author} {\bibfnamefont {B.~R.}\ \bibnamefont {Noack}},\ }\bibfield  {title} {\enquote {\bibinfo {title} {Dynamic feature-based deep reinforcement learning for flow control of circular cylinder with sparse surface pressure sensing},}\ }\href@noop {} {\bibfield  {journal} {\bibinfo  {journal} {Journal of fluid mechanics}\ }\textbf {\bibinfo {volume} {988}},\ \bibinfo {pages} {A4} (\bibinfo {year} {2024})}\BibitemShut {NoStop}%
\bibitem [{\citenamefont {Devlin}\ \emph {et~al.}(2019)\citenamefont {Devlin}, \citenamefont {Chang}, \citenamefont {Lee},\ and\ \citenamefont {Toutanova}}]{DevlinJacob2019BPoD}%
  \BibitemOpen
  \bibfield  {author} {\bibinfo {author} {\bibfnamefont {J.}~\bibnamefont {Devlin}}, \bibinfo {author} {\bibfnamefont {M.-W.}\ \bibnamefont {Chang}}, \bibinfo {author} {\bibfnamefont {K.}~\bibnamefont {Lee}}, \ and\ \bibinfo {author} {\bibfnamefont {K.}~\bibnamefont {Toutanova}},\ }\href {https://arxiv.org/abs/1810.04805} {\enquote {\bibinfo {title} {Bert: Pre-training of deep bidirectional transformers for language understanding},}\ } (\bibinfo {year} {2019}),\ \Eprint {http://arxiv.org/abs/1810.04805} {arXiv:1810.04805 [cs.CL]} \BibitemShut {NoStop}%
\bibitem [{\citenamefont {Li}\ \emph {et~al.}(2023)\citenamefont {Li}, \citenamefont {Li}, \citenamefont {Savarese},\ and\ \citenamefont {Hoi}}]{LiJunnan2023BBLP}%
  \BibitemOpen
  \bibfield  {author} {\bibinfo {author} {\bibfnamefont {J.}~\bibnamefont {Li}}, \bibinfo {author} {\bibfnamefont {D.}~\bibnamefont {Li}}, \bibinfo {author} {\bibfnamefont {S.}~\bibnamefont {Savarese}}, \ and\ \bibinfo {author} {\bibfnamefont {S.}~\bibnamefont {Hoi}},\ }\href {https://arxiv.org/abs/2301.12597} {\enquote {\bibinfo {title} {Blip-2: Bootstrapping language-image pre-training with frozen image encoders and large language models},}\ } (\bibinfo {year} {2023}),\ \Eprint {http://arxiv.org/abs/2301.12597} {arXiv:2301.12597 [cs.CV]} \BibitemShut {NoStop}%
\bibitem [{\citenamefont {Chen}\ \emph {et~al.}(2020)\citenamefont {Chen}, \citenamefont {Kornblith}, \citenamefont {Norouzi},\ and\ \citenamefont {Hinton}}]{ChenTing2020ASFf}%
  \BibitemOpen
  \bibfield  {author} {\bibinfo {author} {\bibfnamefont {T.}~\bibnamefont {Chen}}, \bibinfo {author} {\bibfnamefont {S.}~\bibnamefont {Kornblith}}, \bibinfo {author} {\bibfnamefont {M.}~\bibnamefont {Norouzi}}, \ and\ \bibinfo {author} {\bibfnamefont {G.}~\bibnamefont {Hinton}},\ }\href {https://arxiv.org/abs/2002.05709} {\enquote {\bibinfo {title} {A simple framework for contrastive learning of visual representations},}\ } (\bibinfo {year} {2020}),\ \Eprint {http://arxiv.org/abs/2002.05709} {arXiv:2002.05709} \BibitemShut {NoStop}%
\bibitem [{\citenamefont {He}\ \emph {et~al.}(2022)\citenamefont {He}, \citenamefont {Chen}, \citenamefont {Xie}, \citenamefont {Li}, \citenamefont {Doll{\'a}r},\ and\ \citenamefont {Girshick}}]{he2022masked}%
  \BibitemOpen
  \bibfield  {author} {\bibinfo {author} {\bibfnamefont {K.}~\bibnamefont {He}}, \bibinfo {author} {\bibfnamefont {X.}~\bibnamefont {Chen}}, \bibinfo {author} {\bibfnamefont {S.}~\bibnamefont {Xie}}, \bibinfo {author} {\bibfnamefont {Y.}~\bibnamefont {Li}}, \bibinfo {author} {\bibfnamefont {P.}~\bibnamefont {Doll{\'a}r}}, \ and\ \bibinfo {author} {\bibfnamefont {R.}~\bibnamefont {Girshick}},\ }\bibfield  {title} {\enquote {\bibinfo {title} {Masked autoencoders are scalable vision learners},}\ }in\ \href@noop {} {\emph {\bibinfo {booktitle} {Proceedings of the IEEE/CVF conference on computer vision and pattern recognition}}}\ (\bibinfo {year} {2022})\ pp.\ \bibinfo {pages} {16000--16009}\BibitemShut {NoStop}%
\bibitem [{\citenamefont {Gao}\ \emph {et~al.}(2024)\citenamefont {Gao}, \citenamefont {Fu}, \citenamefont {Wang}, \citenamefont {Qian}, \citenamefont {Feng},\ and\ \citenamefont {Fu}}]{gao2024mind3dreconstructhighquality3d}%
  \BibitemOpen
  \bibfield  {author} {\bibinfo {author} {\bibfnamefont {J.}~\bibnamefont {Gao}}, \bibinfo {author} {\bibfnamefont {Y.}~\bibnamefont {Fu}}, \bibinfo {author} {\bibfnamefont {Y.}~\bibnamefont {Wang}}, \bibinfo {author} {\bibfnamefont {X.}~\bibnamefont {Qian}}, \bibinfo {author} {\bibfnamefont {J.}~\bibnamefont {Feng}}, \ and\ \bibinfo {author} {\bibfnamefont {Y.}~\bibnamefont {Fu}},\ }\href {https://arxiv.org/abs/2312.07485} {\enquote {\bibinfo {title} {Mind-3d: Reconstruct high-quality 3d objects in human brain},}\ } (\bibinfo {year} {2024}),\ \Eprint {http://arxiv.org/abs/2312.07485} {arXiv:2312.07485 [cs.CV]} \BibitemShut {NoStop}%
\bibitem [{\citenamefont {Srinivas}, \citenamefont {Laskin},\ and\ \citenamefont {Abbeel}(2020)}]{SrinivasAravind2020CCUR}%
  \BibitemOpen
  \bibfield  {author} {\bibinfo {author} {\bibfnamefont {A.}~\bibnamefont {Srinivas}}, \bibinfo {author} {\bibfnamefont {M.}~\bibnamefont {Laskin}}, \ and\ \bibinfo {author} {\bibfnamefont {P.}~\bibnamefont {Abbeel}},\ }\href {https://arxiv.org/abs/2004.04136} {\enquote {\bibinfo {title} {Curl: Contrastive unsupervised representations for reinforcement learning},}\ } (\bibinfo {year} {2020}),\ \Eprint {http://arxiv.org/abs/2004.04136} {arXiv:2004.04136 [cs.LG]} \BibitemShut {NoStop}%
\bibitem [{\citenamefont {Stooke}\ \emph {et~al.}(2021)\citenamefont {Stooke}, \citenamefont {Lee}, \citenamefont {Abbeel},\ and\ \citenamefont {Laskin}}]{StookeAdam2021DRLf}%
  \BibitemOpen
  \bibfield  {author} {\bibinfo {author} {\bibfnamefont {A.}~\bibnamefont {Stooke}}, \bibinfo {author} {\bibfnamefont {K.}~\bibnamefont {Lee}}, \bibinfo {author} {\bibfnamefont {P.}~\bibnamefont {Abbeel}}, \ and\ \bibinfo {author} {\bibfnamefont {M.}~\bibnamefont {Laskin}},\ }\bibfield  {title} {\enquote {\bibinfo {title} {Decoupling representation learning from reinforcement learning},}\ }in\ \href@noop {} {\emph {\bibinfo {booktitle} {Proceedings of Machine Learning Research}}},\ Vol.\ \bibinfo {volume} {139}\ (\bibinfo {year} {2021})\ pp.\ \bibinfo {pages} {9870--9879}\BibitemShut {NoStop}%
\bibitem [{\citenamefont {Fukami}, \citenamefont {Fukagata},\ and\ \citenamefont {Taira}(2019)}]{FukamiKai2019Srot}%
  \BibitemOpen
  \bibfield  {author} {\bibinfo {author} {\bibfnamefont {K.}~\bibnamefont {Fukami}}, \bibinfo {author} {\bibfnamefont {K.}~\bibnamefont {Fukagata}}, \ and\ \bibinfo {author} {\bibfnamefont {K.}~\bibnamefont {Taira}},\ }\bibfield  {title} {\enquote {\bibinfo {title} {Super-resolution reconstruction of turbulent flows with machine learning},}\ }\href@noop {} {\bibfield  {journal} {\bibinfo  {journal} {Journal of fluid mechanics}\ }\textbf {\bibinfo {volume} {870}},\ \bibinfo {pages} {106--120} (\bibinfo {year} {2019})}\BibitemShut {NoStop}%
\bibitem [{\citenamefont {Murata}, \citenamefont {Fukami},\ and\ \citenamefont {Fukagata}(2020)}]{MurataTakaaki2020Nmdw}%
  \BibitemOpen
  \bibfield  {author} {\bibinfo {author} {\bibfnamefont {T.}~\bibnamefont {Murata}}, \bibinfo {author} {\bibfnamefont {K.}~\bibnamefont {Fukami}}, \ and\ \bibinfo {author} {\bibfnamefont {K.}~\bibnamefont {Fukagata}},\ }\bibfield  {title} {\enquote {\bibinfo {title} {Nonlinear mode decomposition with convolutional neural networks for fluid dynamics},}\ }\href@noop {} {\bibfield  {journal} {\bibinfo  {journal} {Journal of fluid mechanics}\ }\textbf {\bibinfo {volume} {882}},\ \bibinfo {pages} {A13} (\bibinfo {year} {2020})}\BibitemShut {NoStop}%
\bibitem [{\citenamefont {Racca}, \citenamefont {Doan},\ and\ \citenamefont {Magri}(2023)}]{RaccaAlberto2023Ptdw}%
  \BibitemOpen
  \bibfield  {author} {\bibinfo {author} {\bibfnamefont {A.}~\bibnamefont {Racca}}, \bibinfo {author} {\bibfnamefont {N.~A.~K.}\ \bibnamefont {Doan}}, \ and\ \bibinfo {author} {\bibfnamefont {L.}~\bibnamefont {Magri}},\ }\bibfield  {title} {\enquote {\bibinfo {title} {Predicting turbulent dynamics with the convolutional autoencoder echo state network},}\ }\href@noop {} {\bibfield  {journal} {\bibinfo  {journal} {Journal of fluid mechanics}\ }\textbf {\bibinfo {volume} {975}},\ \bibinfo {pages} {A2} (\bibinfo {year} {2023})}\BibitemShut {NoStop}%
\bibitem [{\citenamefont {Cho}\ \emph {et~al.}(2014)\citenamefont {Cho}, \citenamefont {van Merrienboer}, \citenamefont {Gulcehre}, \citenamefont {Bahdanau}, \citenamefont {Bougares}, \citenamefont {Schwenk},\ and\ \citenamefont {Bengio}}]{ChoKyunghyun2014LPRu}%
  \BibitemOpen
  \bibfield  {author} {\bibinfo {author} {\bibfnamefont {K.}~\bibnamefont {Cho}}, \bibinfo {author} {\bibfnamefont {B.}~\bibnamefont {van Merrienboer}}, \bibinfo {author} {\bibfnamefont {C.}~\bibnamefont {Gulcehre}}, \bibinfo {author} {\bibfnamefont {D.}~\bibnamefont {Bahdanau}}, \bibinfo {author} {\bibfnamefont {F.}~\bibnamefont {Bougares}}, \bibinfo {author} {\bibfnamefont {H.}~\bibnamefont {Schwenk}}, \ and\ \bibinfo {author} {\bibfnamefont {Y.}~\bibnamefont {Bengio}},\ }\href {https://arxiv.org/abs/1406.1078} {\enquote {\bibinfo {title} {Learning phrase representations using rnn encoder-decoder for statistical machine translation},}\ } (\bibinfo {year} {2014}),\ \Eprint {http://arxiv.org/abs/1406.1078} {arXiv:1406.1078 [cs.CL]} \BibitemShut {NoStop}%
\bibitem [{\citenamefont {van~den Oord}, \citenamefont {Li},\ and\ \citenamefont {Vinyals}(2019)}]{vandenOordAaron2019RLwC}%
  \BibitemOpen
  \bibfield  {author} {\bibinfo {author} {\bibfnamefont {A.}~\bibnamefont {van~den Oord}}, \bibinfo {author} {\bibfnamefont {Y.}~\bibnamefont {Li}}, \ and\ \bibinfo {author} {\bibfnamefont {O.}~\bibnamefont {Vinyals}},\ }\href {https://arxiv.org/abs/1807.03748} {\enquote {\bibinfo {title} {Representation learning with contrastive predictive coding},}\ } (\bibinfo {year} {2019}),\ \Eprint {http://arxiv.org/abs/1807.03748} {arXiv:1807.03748 [cs.LG]} \BibitemShut {NoStop}%
\bibitem [{\citenamefont {Schulman}\ \emph {et~al.}(2017)\citenamefont {Schulman}, \citenamefont {Wolski}, \citenamefont {Dhariwal}, \citenamefont {Radford},\ and\ \citenamefont {Klimov}}]{SchulmanJohn2017PPOA}%
  \BibitemOpen
  \bibfield  {author} {\bibinfo {author} {\bibfnamefont {J.}~\bibnamefont {Schulman}}, \bibinfo {author} {\bibfnamefont {F.}~\bibnamefont {Wolski}}, \bibinfo {author} {\bibfnamefont {P.}~\bibnamefont {Dhariwal}}, \bibinfo {author} {\bibfnamefont {A.}~\bibnamefont {Radford}}, \ and\ \bibinfo {author} {\bibfnamefont {O.}~\bibnamefont {Klimov}},\ }\href {https://arxiv.org/abs/1707.06347} {\enquote {\bibinfo {title} {Proximal policy optimization algorithms},}\ } (\bibinfo {year} {2017}),\ \Eprint {http://arxiv.org/abs/1707.06347} {arXiv:1707.06347 [cs.LG]} \BibitemShut {NoStop}%
\bibitem [{\citenamefont {Shu}, \citenamefont {Liu},\ and\ \citenamefont {Chew}(2007)}]{ShuC.2007Anib}%
  \BibitemOpen
  \bibfield  {author} {\bibinfo {author} {\bibfnamefont {C.}~\bibnamefont {Shu}}, \bibinfo {author} {\bibfnamefont {N.}~\bibnamefont {Liu}}, \ and\ \bibinfo {author} {\bibfnamefont {Y.}~\bibnamefont {Chew}},\ }\bibfield  {title} {\enquote {\bibinfo {title} {A novel immersed boundary velocity correction–lattice boltzmann method and its application to simulate flow past a circular cylinder},}\ }\href@noop {} {\bibfield  {journal} {\bibinfo  {journal} {Journal of computational physics}\ }\textbf {\bibinfo {volume} {226}},\ \bibinfo {pages} {1607--1622} (\bibinfo {year} {2007})}\BibitemShut {NoStop}%
\bibitem [{\citenamefont {Liu}, \citenamefont {Zheng},\ and\ \citenamefont {Sung}(1998)}]{LIU199835}%
  \BibitemOpen
  \bibfield  {author} {\bibinfo {author} {\bibfnamefont {C.}~\bibnamefont {Liu}}, \bibinfo {author} {\bibfnamefont {X.}~\bibnamefont {Zheng}}, \ and\ \bibinfo {author} {\bibfnamefont {C.}~\bibnamefont {Sung}},\ }\bibfield  {title} {\enquote {\bibinfo {title} {Preconditioned multigrid methods for unsteady incompressible flows},}\ }\href@noop {} {\bibfield  {journal} {\bibinfo  {journal} {Journal of Computational Physics}\ }\textbf {\bibinfo {volume} {139}},\ \bibinfo {pages} {35--57} (\bibinfo {year} {1998})}\BibitemShut {NoStop}%
\bibitem [{\citenamefont {Su}, \citenamefont {Lai},\ and\ \citenamefont {Lin}(2007)}]{SuShen-Wei2007Aibt}%
  \BibitemOpen
  \bibfield  {author} {\bibinfo {author} {\bibfnamefont {S.-W.}\ \bibnamefont {Su}}, \bibinfo {author} {\bibfnamefont {M.-C.}\ \bibnamefont {Lai}}, \ and\ \bibinfo {author} {\bibfnamefont {C.-A.}\ \bibnamefont {Lin}},\ }\bibfield  {title} {\enquote {\bibinfo {title} {An immersed boundary technique for simulating complex flows with rigid boundary},}\ }\href@noop {} {\bibfield  {journal} {\bibinfo  {journal} {Computers \& fluids}\ }\textbf {\bibinfo {volume} {36}},\ \bibinfo {pages} {313--324} (\bibinfo {year} {2007})}\BibitemShut {NoStop}%
\bibitem [{\citenamefont {Tseng}\ and\ \citenamefont {Ferziger}(2003)}]{TsengYu-Heng2003Agib}%
  \BibitemOpen
  \bibfield  {author} {\bibinfo {author} {\bibfnamefont {Y.-H.}\ \bibnamefont {Tseng}}\ and\ \bibinfo {author} {\bibfnamefont {J.~H.}\ \bibnamefont {Ferziger}},\ }\bibfield  {title} {\enquote {\bibinfo {title} {A ghost-cell immersed boundary method for flow in complex geometry},}\ }\href@noop {} {\bibfield  {journal} {\bibinfo  {journal} {Journal of computational physics}\ }\textbf {\bibinfo {volume} {192}},\ \bibinfo {pages} {593--623} (\bibinfo {year} {2003})}\BibitemShut {NoStop}%
\bibitem [{\citenamefont {Williamson}(1988)}]{WilliamsonC.H.K.1988Daua}%
  \BibitemOpen
  \bibfield  {author} {\bibinfo {author} {\bibfnamefont {C.~H.~K.}\ \bibnamefont {Williamson}},\ }\bibfield  {title} {\enquote {\bibinfo {title} {Defining a universal and continuous strouhal–reynolds number relationship for the laminar vortex shedding of a circular cylinder},}\ }\href@noop {} {\bibfield  {journal} {\bibinfo  {journal} {The Physics of fluids (1958)}\ }\textbf {\bibinfo {volume} {31}},\ \bibinfo {pages} {2742--2744} (\bibinfo {year} {1988})}\BibitemShut {NoStop}%
\bibitem [{\citenamefont {Hurlbut}, \citenamefont {Spaulding},\ and\ \citenamefont {White}(1982)}]{HurlbutS.E1982NSfL}%
  \BibitemOpen
  \bibfield  {author} {\bibinfo {author} {\bibfnamefont {S.~E.}\ \bibnamefont {Hurlbut}}, \bibinfo {author} {\bibfnamefont {M.~L.}\ \bibnamefont {Spaulding}}, \ and\ \bibinfo {author} {\bibfnamefont {F.~M.}\ \bibnamefont {White}},\ }\bibfield  {title} {\enquote {\bibinfo {title} {Numerical solution for laminar two dimensional flow about a cylinder oscillating in a uniform stream},}\ }\href@noop {} {\bibfield  {journal} {\bibinfo  {journal} {Journal of fluids engineering}\ }\textbf {\bibinfo {volume} {104}},\ \bibinfo {pages} {214--220} (\bibinfo {year} {1982})}\BibitemShut {NoStop}%
\end{thebibliography}%

\end{document}